\documentclass[journal]{IEEEtran}
\usepackage{cite}
\ifCLASSINFOpdf
\else
\fi

\usepackage{amsmath,amssymb,bm,graphicx,color}
\usepackage{url}
\begin{document}
\title{When Gaussian Process Meets Big Data: A Review of Scalable GPs}

\author{Haitao~Liu,
	Yew-Soon~Ong,~\IEEEmembership{Fellow,~IEEE,}
	Xiaobo~Shen,	
	and~Jianfei~Cai,~\IEEEmembership{Senior~Member,~IEEE}% <-this % stops a space
\thanks{Haitao Liu is with the Rolls-Royce@NTU Corporate Lab, Nanyang Technological University, Singapore, 637460. E-mail: htliu@ntu.edu.sg}% <-this % stops a space
\thanks{Yew-Soon Ong, Xiaobo Shen and Jianfei Cai are with School of Computer Science and Engineering, Nanyang Technological University, Singapore, 639798. E-mail: \{asysong, xbshen, asjfcai\}@ntu.edu.sg.}% <-this % stops a space
%\thanks{Manuscript received xxx, 2018; revised xxx, 2018.}
}


\markboth{IEEE}%
%\markboth{}%
{Shell \MakeLowercase{\textit{et al.}}: Bare Demo of IEEEtran.cls for IEEE Journals}



\maketitle

\begin{abstract}
The vast quantity of information brought by big data as well as the evolving computer hardware encourages success stories in the machine learning community. In the meanwhile, it poses challenges for the Gaussian process (GP) regression, a well-known non-parametric and interpretable Bayesian model, which suffers from cubic complexity to data size. To improve the scalability while retaining desirable prediction quality, a variety of scalable GPs have been presented. But they have not yet been comprehensively reviewed and analyzed in order to be well understood by both academia and industry. The review of scalable GPs in the GP community is timely and important due to the explosion of data size. To this end, this paper is devoted to the review on state-of-the-art scalable GPs involving two main categories: global approximations which distillate the entire data and local approximations which divide the data for subspace learning. Particularly, for global approximations, we mainly focus on sparse approximations comprising prior approximations which modify the prior but perform exact inference, posterior approximations which retain exact prior but perform approximate inference, and structured sparse approximations which exploit specific structures in kernel matrix; for local approximations, we highlight the mixture/product of experts that conducts model averaging from multiple local experts to boost predictions. To present a complete review, recent advances for improving the scalability and capability of scalable GPs are reviewed. Finally, the extensions and open issues regarding the implementation of scalable GPs in various scenarios are reviewed and discussed to inspire novel ideas for future research avenues.
\end{abstract}


% \fi
\IEEEpeerreviewmaketitle

\section{Introduction}
\IEEEPARstart{I}{n} the era of big data, the vast quantity of information poses the demand of effective and efficient analysis, interpretation and prediction to explore the benefits lie ahead. Thanks to the big data, the machine learning community tells many success stories~\cite{hensman2013gaussian, del2014use, lecun2015deep, silver2017mastering} while still leaving many challenges. We focus on Gaussian process (GP) regression~\cite{rasmussen2006gaussian}, also known as Kriging in geostatistics~\cite{matheron1963principles}, and surrogates or emulators in computer experiments~\cite{sacks1989design}. The GP is a non-parametric statistical model which has been extensively used in various scenarios, e.g., active learning~\cite{liu2018survey}, multi-task learning~\cite{alvarez2012kernels, liu2018remarks}, manifold learning~\cite{lawrence2005probabilistic}, and optimization~\cite{shahriari2016taking}.

\textit{Big data} in the GP community mainly refers to one of the 5V challenges~\cite{yin2015big}: the \textit{volume} which represents the huge amount of data points to be stored, processed and analyzed, incurring high computational complexity for current GP paradigms. It is worth noting that this review mainly focuses on scalable GPs for large-scale regression but not on all forms of GPs or other machine learning models.

Given $n$ training points $\bm{X} = \{\bm{x}_i \in R^d\}_{i=1}^n$ and their observations $\bm{y} = \{y_i = y(\bm{x}_i) \in R\}_{i=1}^n$, GP seeks to infer the latent function $f: R^d \mapsto R$ in the function space $\mathcal{GP}(m(\bm{x}), k(\bm{x}, \bm{x}'))$ defined by the mean $m(.)$ and the kernel $k(.,.)$. The most prominent weakness of standard GP is that it suffers from a cubic time complexity $\mathcal{O}(n^3)$ because of the inversion and determinant of the $n \times n$ kernel matrix $\bm{K}_{nn} = k(\bm{X}, \bm{X})$. This limits the scalability of GP and and makes it unaffordable for large-scale datasets.

\begin{figure}[!htb] 
	\centering
	\includegraphics[width=0.42\textwidth]{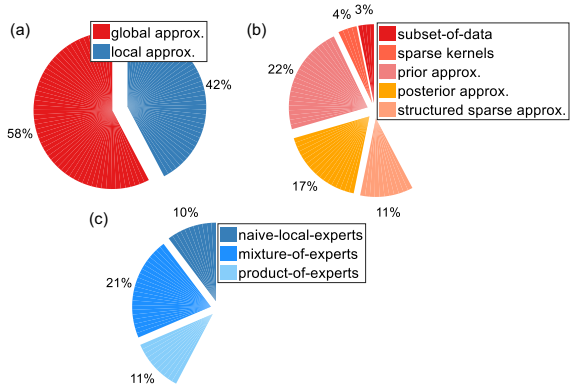}
	\caption{Percentages of the categories for (a) scalable GPs including (b) global approximations and (c) local approximations in the literature surveyed.}
	\label{Fig_global_local_percentage} 
\end{figure}

Hence, scalable GPs devote to \textit{improving the scalability of full GP while retaining favorable prediction quality} for big data. The extensive literature review summarized in Fig.~\ref{Fig_global_local_percentage} classifies scalable GPs into two main categories including

(a) \textit{Global approximations} which approximate the kernel matrix $\bm{K}_{nn}$ through global distillation. The distillation can be achieved by (i) a subset of the training data with $m$ ($m \ll n$) points (subset-of-data~\cite{chalupka2013framework}), resulting in a smaller kernel matrix $\bm{K}_{mm}$; (ii) the remove of uncorrelated entries in $\bm{K}_{nn}$ (sparse kernels~\cite{gneiting2002compactly}), resulting in a sparse kernel matrix $\tilde{\bm{K}}_{nn}$ with many zero entries; and (iii) the low-rank representation measured between $m$ inducing points and $n$ training points (sparse approximations~\cite{quinonero2005unifying, titsias2009variational, hensman2013gaussian, wilson2015kernel}), resulting in the Nystr\"{o}m approximation $\bm{K}_{nn} \approx \bm{K}_{nm} \bm{K}^{-1}_{mm} \bm{K}_{mn}$.

(b) \textit{Local approximations} which follow the \textit{divide-and-conquer} (D\&C) idea to focus on the local subsets of training data. Efficiently, local approximations only need to tackle a \textit{local} expert with $m_0$ ($m_0 \ll n$) data points at each time~\cite{gramacy2008bayesian, gramacy2016lagp}. Additionally, to produce smooth predictions equipped with valid uncertainty, modeling averaging has been employed through mixture or product of experts~\cite{yuksel2012twenty, masoudnia2014mixture, rasmussen2002infinite, sun2011variational, hinton2002training, deisenroth2015distributed, rulliere2018nested, liu2018generalized}.

\begin{figure}[!htb] 
	\centering
	\includegraphics[width=0.42\textwidth]{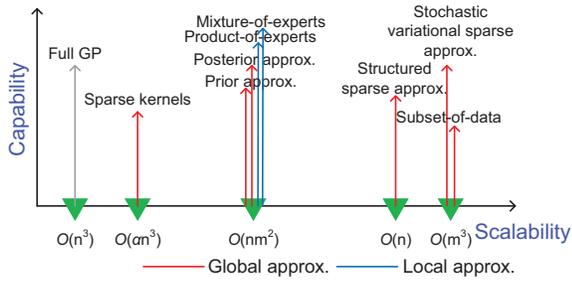}
	\caption{Comparison of scalable GPs regarding scalability and model capability, where $0 < \alpha <1$; $m$ is the inducing size for sparse approximations, and the subset size for subset-of-data and local approximations.}
	\label{Fig_complexity_scalableGPs} 
\end{figure}

As depicted in Fig.~\ref{Fig_complexity_scalableGPs}, in terms of scalability, most of the sparse approximations using $m$ inducing points and the local approximations using $m_0=m$ data points for each expert have the same training complexity as $\mathcal{O}(nm^2)$, and they can be further sped up through parallel/distributed computing~\cite{gal2014distributed, dai2014gaussian, matthews2017gpflow, gramacy2016lagp, gramacy2016speeding, gramacy2014massively}. When organizing the inducing points into Kronecker structure, sparse approximations can further reduce the complexity to $\mathcal{O}(n)$~\cite{wilson2015kernel, wilson2016deep}. In the meantime, by reorganizing the variational lower bound, stochastic optimization is available for sparse approximations with a remarkable complexity of $\mathcal{O}(m^3)$~\cite{hensman2013gaussian, hoang2015unifying, peng2017asynchronous}, enabling the regression with million- and even billion-sized data points~\cite{rivera2017forecasting, peng2017asynchronous}.

It is notable that we welcome GPs with high scalability but require producing favorable predictions, i.e., good model capability. For example, though showing a remarkable complexity of $\mathcal{O}(m^3)$, we cannot expect the subset-of-data to perform well with increasing $n$. In terms of model capability, global approximations are capable of capturing the global patterns (long-term spatial correlations) but often filter out the local patterns due to the limited global inducing set. In contrast, due to the local nature, local approximations favor capturing local patterns (non-stationary features), enabling them to outperform global approximations for complicated tasks, see the \textit{solar} example in~\cite{liu2019understanding}. The drawback however is that they ignore the global patterns to risk discontinuous predictions and local over-fitting. Recently, attempts have been made to improve the model capability through, for example, the inter-domain strategy~\cite{lazaro2009inter}, hierarchical structure~\cite{lee2017hierarchically}, and hybrid of global \& local approximations or neural networks (NNs)~\cite{snelson2007local, nguyen2014fast, wilson2016deep}, showcasing the state-of-the-art performance~\cite{wilson2016deep, hoang2016distributed}.

\begin{figure*}[!htb] 
	\centering
	\includegraphics[width=0.6\textwidth]{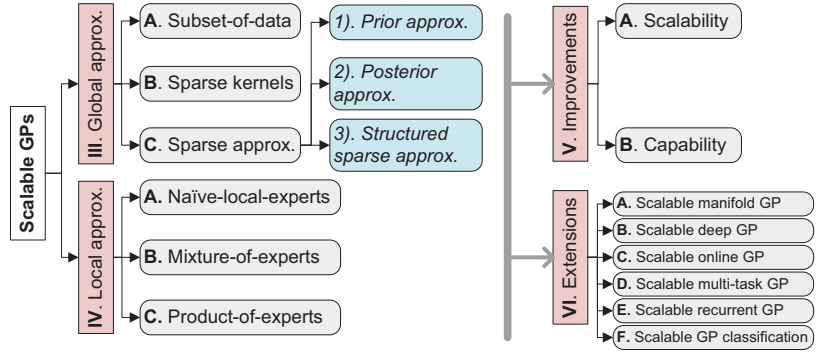}
	\caption{A skeletal overview of scalable GPs.}
	\label{Fig_skeleton_survey} 
\end{figure*}

The development and success of scalable GPs pose the demand of comprehensive review including the methodological characteristics and comparisons for better understanding. To the best of our knowledge, a detailed survey on various scalable GPs for large-scale regression has not been conducted in the literature before and such a work in the GP community is timely and important due to the explosion of data size.\footnote{The survey~\cite{quinonero2005unifying} at 14 years ago focuses on the prior approximations, which is just a part of our review in section~\ref{sec_prior_approx}. The recent comparison and survey~\cite{camps2016survey, rivera2017forecasting} provide however a quick and rough review without detailed analysis.}

We thus consider a skeletal overview in Fig.~\ref{Fig_skeleton_survey} to classify, review and analyze state-of-the-art scalable GPs. Specifically, with a quick introduction of standard GP regression in section~\ref{sec_exactGP}, the two main categories of scalable GPs, global and local approximations, are then comprehensively reviewed in sections~\ref{sec_global_approx} and~\ref{sec_local_approx}. Moreover, section~\ref{sec_global_local} reviews the improvements for scalable GPs in terms of scalability and capability. Thereafter, section~\ref{sec_futurework} discusses the extensions of scalable GPs in different scenarios to highlight potential research avenues. Finally, section~\ref{sec_conclusions} offers concluding remarks.

\section{Gaussian process regression revisited}
\label{sec_exactGP}
The non-parametric GP regression (GPR) places a GP prior over the latent function as $f(\bm{x}) \sim \mathcal{GP}(m(\bm{x}), k(\bm{x}, \bm{x}'))$~\cite{rasmussen2006gaussian}.
The mean function $m(\bm{x})$ is often taken as zero. The kernel function $k(\bm{x}, \bm{x}')$ controls the smoothness of GP and is often taken as the squared exponential (SE) function equipped with automatic relevance determination (ARD)
\begin{equation} \label{eq_SE}
k_{\mathrm{SE}}(\bm{x}, \bm{x}') = \sigma^2_f \exp(-0.5 (\bm{x} - \bm{x}')^{\mathsf{T}} \bm{\Delta}^{-1} (\bm{x} - \bm{x}')),
\end{equation}
where $\bm{\Delta} = \mathrm{diag}[l_1^2, \cdots, l_d^2]$ comprises the length-scales along $d$ dimensions, and $\sigma^2_f$ is the signal variance. For other conventional kernels, e.g., the Mat\'ern kernel, please refer to~\cite{rasmussen2006gaussian}.

Given the training data $\mathcal{D} = \{\bm{X}, \bm{y} \}$ where $y(\bm{x}_i) = f(\bm{x}_i) + \epsilon$ with the \textit{iid} noise $\epsilon \sim \mathcal{N}(0, \sigma^2_{\epsilon})$, we obtain the model evidence (marginal likelihood) $p(\bm{y}| \bm{\theta}) = \int p(\bm{y}|\bm{f}) p(\bm{f}) d\bm{f} = \mathcal{N}(\bm{y}|\bm{0}, \bm{K}_{nn}^{\epsilon})$,\footnote{For the sake of clarity, the hyperparameters $\bm{\theta}$ below are omitted from the conditioning of the distribution.} 
where $\bm{K}_{nn}^{\epsilon} = \bm{K}_{nn} + \sigma^2_{\epsilon} \bm{I}_n$, and $\bm{\theta}$ comprises the hyperparameters which could be inferred by maximizing
\begin{equation} \label{eq_logP(y)_GP}
\log p(\bm{y}) = - \frac{n}{2} \log2\pi - \frac{1}{2} \log |\bm{K}_{nn}^{\epsilon}| - \frac{1}{2} \bm{y}^{\mathsf{T}} (\bm{K}_{nn}^{\epsilon})^{-1} \bm{y},
\end{equation}
which automatically achieves the \textit{bias-variance} trade-off.

Thereafter, the predictive distribution $p(f_*|\mathcal{D}, \bm{x}_*) = \mathcal{N}(f_*|\mu(\bm{x}_*), \sigma^2_*(\bm{x}_*))$ at a test point $\bm{x}_*$ has the mean and variance respectively expressed as
\begin{subequations}
	\label{eq_GP_pred}
	\begin{align}
	\mu(\bm{x}_*) =& \bm{k}_{*n} (\bm{K}_{nn}^{\epsilon})^{-1} \bm{y}, \\
	\sigma^2(\bm{x}_*) =& k_{**} - \bm{k}_{*n} (\bm{K}_{nn}^{\epsilon})^{-1} \bm{k}_{n*},
	\end{align}
\end{subequations}
where $\bm{k}_{*n} = k(\bm{x}_*, \bm{X})$ and $k_{**} = k(\bm{x}_*, \bm{x}_*)$. For $y_*$, we need to consider the noise such that $p(y_*|\mathcal{D}, \bm{x}_*) = \mathcal{N}(y_*|\mu(\bm{x}_*), \sigma^2_*(\bm{x}_*)+\sigma^2_{\epsilon})$.

Alternatively, we can interpret the GP from the \textit{weight-space} view as an extension of the Bayesian linear model as
\begin{equation} \label{eq_weight_GP}
f(\bm{x}) = \bm{\phi}(\bm{x})^{\mathsf{T}} \bm{w}, \quad y(\bm{x}) = f(\bm{x}) + \epsilon,
\end{equation}
where the Gaussian prior is placed on the weights as $p(\bm{w}) = \mathcal{N}(\bm{w}|\bm{0}, \bm{\Sigma})$; $\bm{\phi}(\bm{x}) = [\phi_1(\bm{x}), \cdots, \phi_v(\bm{x})]^{\mathsf{T}}$ maps the $d$-dimensional input $\bm{x}$ into a $v$-dimensional feature space. Equivalently, we derive the kernel as $k(\bm{x}, \bm{x}') = \bm{\phi}(\bm{x})^{\mathsf{T}} \bm{\Sigma} \bm{\phi}(\bm{x}')$. Particularly, the SE kernel~\eqref{eq_SE} can  be recovered from an infinite number ($v \to \infty$) of Gaussian-shaped basis functions $\{ \phi_c(\bm{x}) \}_{c=1}^v$ centered everywhere. 

The computational bottleneck of GP inference in~\eqref{eq_logP(y)_GP} is solving the linear system $(\bm{K}_{nn}^{\epsilon})^{-1} \bm{y}$ and the determinant $|\bm{K}_{nn}^{\epsilon}|$. Traditionally, we use the $\mathcal{O}(n^3)$ Cholesky decomposition $\bm{K}_{nn}^{\epsilon} = \bm{L} \bm{L}^{\mathsf{T}}$ such that $(\bm{K}_{nn}^{\epsilon})^{-1} \bm{y} = \bm{L}^{\mathsf{T}} \setminus (\bm{L} \setminus \bm{y})$ and $\log |\bm{K}_{nn}^{\epsilon}| = 2 \sum_{i=1}^n \log \bm{L}_{ii}$. As for predictions in~\eqref{eq_GP_pred}, the mean costs $\mathcal{O}(n)$ and the variance costs $\mathcal{O}(n^2)$ per test case through pre-computations.

In order to improve the scalability of standard GP for big data, the scalable GPs have been extensively presented and studied in recent years. In what follows, we classify current scalable GPs into global approximations and local approximations, and comprehensively analyze them to showcase their methodological characteristics.

\section{Global approximations}
\label{sec_global_approx}
Global approximations achieve the sparsity of the full kernel matrix $\bm{K}_{nn}$, which is crucial for scalability, through (i) using a subset of the training data (subset-of-data); (ii) removing the entries of $\bm{K}_{nn}$ with low correlations (sparse kernels); and (iii) employing a low-rank representation (sparse approximations).

\subsection{Subset-of-data}
Subset-of-data (SoD) is the simplest strategy to approximate the full GP by using a subset $\mathcal{D}_{\mathrm{sod}}$ of the training data $\mathcal{D}$. Hence, the SoD retains the standard GP inference at lower time complexity of $\mathcal{O}(m^3)$, since it operates on $\bm{K}_{mm}$ which only comprises $m$ ($m \ll n$) data points. A recent theoretical work~\cite{hayashi2019random} analyzes the error bounds for the prediction and generalization of SoD through a graphon-based framework, indicating a better speed-accuracy trade-off in comparison to other approximations reviewed below when $n$ is sufficiently large. Though SoD produces reasonable prediction mean for the case with redundant data, it struggles to produce overconfident prediction variance due to the limited subset.

Regarding the selection of $\mathcal{D}_{\mathrm{sod}}$, one could (i) randomly choose $m$ data points from $\mathcal{D}$, (ii) use clustering techniques, e.g., $k$-means and KD tree~\cite{preparata2012computational}, to partition the data into $m$ subsets and choose their centroids as subset points, and (iii) employ active learning criteria, e.g., differential entropy~\cite{herbrich2003fast}, information gain~\cite{seeger2003bayesian} and matching pursuit~\cite{keerthi2006matching}, to sequentially query data points with however higher computing cost.

\subsection{Sparse kernels}
Sparse kernels~\cite{melkumyan2009sparse} attempt to directly achieve a sparse representation $\tilde{\bm{K}}_{nn}$ of $\bm{K}_{nn}$ via the particularly designed compactly supported (CS) kernel, which imposes $k(\bm{x}_i, \bm{x}_j) = 0$ when $|\bm{x}_i - \bm{x}_j|$ exceeds a certain threshold. Therefore, only the non-zero elements in $\tilde{\bm{K}}_{nn}$ are involved in the calculation. As a result, the training complexity of the GP using CS kernel scales as $\mathcal{O}(\alpha n^3)$ with $0 < \alpha < 1$. The main challenge in constructing valid CS kernels is to ensure the positive semi-definite (PSD) of $\tilde{\bm{K}}_{nn}$, i.e., $\bm{v}^{\mathsf{T}} \tilde{\bm{K}}_{nn} \bm{v} \ge 0, \, \forall \bm{v} \in R^n$~\cite{buhmann2001new, gneiting2002compactly, wendland2004scattered, melkumyan2009sparse}. Besides, the GP using CS kernel is potential for capturing local patterns due to the truncation property.

\subsection{Sparse approximations}
\label{sec_sparse_approximations}
Typically, we could conduct eigen-decomposition and choose the first $m$ eigenvalues to approximate the full-rank kernel matrix as $\bm{K}_{nn} \approx \bm{U}_{nm} \bm{\Lambda}_{mm} \bm{U}_{nm}^{\mathsf{T}}$.
Thereafter, it is straightforward to calculate the inversion using the Sherman-Morrison-Woodbury formula
\begin{equation*} \label{eq_woodbury}
\begin{aligned}
(\bm{K}_{nn}^{\epsilon})^{-1} 
\approx \sigma^{-2}_{\epsilon} \bm{I}_n + \sigma^{-2}_{\epsilon} \bm{U}_{nm} (\sigma^{2}_{\epsilon} \bm{\Lambda}_{mm}^{-1} +  \bm{U}_{nm}^{\mathsf{T}} \bm{U}_{nm})^{-1} \bm{U}_{nm}^{\mathsf{T}},
\end{aligned}
\end{equation*}
and the determinant using the Sylvester determinant theorem
\begin{equation*} \label{eq_determinant}
|\bm{K}_{nn}^{\epsilon}| \approx |\bm{\Lambda}_{mm}| |\sigma^{2}_{\epsilon} \bm{\Lambda}_{mm}^{-1} +  \bm{U}_{nm}^{\mathsf{T}} \bm{U}_{nm}|,
\end{equation*} 
resulting in the complexity of $\mathcal{O}(nm^2)$. However, the eigen-decomposition is of limited interest since itself is an $\mathcal{O}(n^3)$ operation. Hence, we approximate the eigen-functions of $\bm{K}_{nn}$ using $m$ data points, leading to the Nystr\"{o}m approximation 
\begin{equation*}
\bm{K}_{nn} \approx \bm{Q}_{nn} = \bm{K}_{nm} \bm{K}_{mm}^{-1} \bm{K}_{nm}^{\mathsf{T}},
\end{equation*}
which greatly improves large-scale kernel learning~\cite{gittens2016revisiting}, and enables naive Nystr\"{o}m GP~\cite{williams2001using}. This scalable GP however may produce negative prediction variances~\cite{williams2002observations}, since (i) it is not a complete generative probabilistic model as the Nystr\"{o}m approximation is only imposed on the training data, and (ii) it cannot guarantee the PSD of kernel matrix.

Inspired by the influential Nystr\"{o}m approximation, sparse approximations build a generative probabilistic model, which achieves the sparsity via $m$ inducing points (also referred to as support points, active set or pseudo points) to optimally summarize the dependency of the whole training data. We introduce a set of inducing pairs $(\bm{X}_m, \bm{f}_m)$. The latent variables $\bm{f}_m$ akin to $\bm{f}$ follow the same GP prior $p(\bm{f}_m) = \mathcal{N}(\bm{0}, \bm{K}_{mm})$. Besides, $\bm{f}_m$ is assumed to be a sufficient statistic for $\bm{f}$, i.e., for any variables $\bm{z}$ it holds $p(\bm{z} | \bm{f}, \bm{f}_m) = p(\bm{z} | \bm{f}_m)$. We could recover the joint prior $p(\bm{f}, f_*)$ by marginalizing out $\bm{f}_m$ as $p(\bm{f}, f_*) = \int p(\bm{f}, f_* | \bm{f}_m) p(\bm{f}_m) d\bm{f}_m$.

In what follows, sparse approximations have three main categories: 
\begin{itemize}
	\item [-] \textit{prior approximations} which approximate the prior but perform exact inference;
	\item [-] \textit{posterior approximations} which retain exact prior but perform approximate inference; and
	\item [-] \textit{structured sparse approximations} which exploit specific structures in kernel matrix.
\end{itemize}

\vspace{1ex}
\subsubsection{\textbf{Prior approximations}}
\label{sec_prior_approx}
Prior approximations~\cite{quinonero2005unifying} modify the joint prior, which is the origin of the cubic complexity, using the independence assumption $\bm{f} \bot f_*|\bm{f}_m$ such that
\begin{equation}
\label{eq_joint_prior}
\begin{aligned}
p(\bm{f}, f_*) &= \int p(\bm{f}| \bm{f}_m) p(f_*|\bm{f}_m) p(\bm{f}_m) d\bm{f}_m,
\end{aligned}
\end{equation} 
where the training and test conditionals write, given a Nystr\"{o}m notation $\bm{Q}_{ab} = \bm{K}_{am} \bm{K}_{mm}^{-1} \bm{K}_{mb}$,
\begin{subequations} \label{eq_training_test_cond}
	\begin{align}
	p(\bm{f} | \bm{f}_m) =& \mathcal{N} (\bm{f}| \bm{K}_{nm} \bm{K}_{mm}^{-1} \bm{f}_m, \bm{K}_{nn} - \bm{Q}_{nn}), \label{eq_training_cond} \\
	p(f_* | \bm{f}_m) =& \mathcal{N} (f_*| \bm{k}_{*m} \bm{K}_{mm}^{-1} \bm{f}_m, k_{**} - Q_{**}). \label{eq_test_cond} 
	\end{align}
\end{subequations}
We see here $\bm{f}_m$ is called inducing variables since the dependencies between $\bm{f}$ and $f_*$ are only induced through $\bm{f}_m$. To obtain computational gains, we modify the training and test conditionals  as
\begin{subequations}
\begin{align}
q(\bm{f} | \bm{f}_m) &= \mathcal{N} (\bm{f}|\bm{K}_{nm} \bm{K}_{mm}^{-1} \bm{f}_m, \tilde{\bm{Q}}_{nn}), \\
q(f_* | \bm{f}_m) &= \mathcal{N} (f_*|\bm{k}_{*m} \bm{K}_{mm}^{-1} \bm{f}_m, \tilde{Q}_{**}).
\end{align}
\end{subequations}
Then, $\log p(\bm{y})$ is approximated by $\log q(\bm{y})$ as
\begin{equation} \label{eq_log_q(y)}
\begin{aligned}
\log q(\bm{y}) = &- \frac{n}{2} \log2\pi - \frac{1}{2} \log |\tilde{\bm{Q}}_{nn} + \bm{Q}_{nn} + \sigma^2_{\epsilon} \bm{I}_n| \\
&- \frac{1}{2} \bm{y}^{\mathsf{T}} (\tilde{\bm{Q}}_{nn} + \bm{Q}_{nn} + \sigma^2_{\epsilon} \bm{I}_n)^{-1} \bm{y}.
\end{aligned}
\end{equation}
It is found that specific selections of $\tilde{\bm{Q}}_{nn}$ enable  calculating $|\tilde{\bm{Q}}_{nn} + \bm{Q}_{nn} + \sigma^2_{\epsilon} \bm{I}_n|$ and $(\tilde{\bm{Q}}_{nn} + \bm{Q}_{nn} + \sigma^2_{\epsilon} \bm{I}_n)^{-1}$ with a substantially reduced complexity of $\mathcal{O}(nm^2)$.

Particularly, the subset-of-regressors (SoR)~\cite{smola2001sparse}, also called deterministic inducing conditional (DIC), imposes deterministic training and test conditionals, i.e., $\tilde{\bm{Q}}_{nn} = \bm{0}$ and $\tilde{Q}_{**} = 0$, as
\begin{subequations} \label{eq_SoR_conditional}
\begin{align}
q_{\mathrm{SoR}}(\bm{f} | \bm{f}_m) &= \mathcal{N} (\bm{f}|\bm{K}_{nm} \bm{K}_{mm}^{-1} \bm{f}_m, \bm{0}), \\ q_{\mathrm{SoR}}(f_* | \bm{f}_m) &= \mathcal{N} (f_*|\bm{k}_{*m} \bm{K}_{mm}^{-1} \bm{f}_m, 0).
\end{align}
\end{subequations}
This is equivalent to applying the Nystr\"{o}m approximation to both training and test data, resulting in a \textit{degenerate}\footnote{It means the kernel $k(.,.)$ has a finite number of non-zero eigenvalues.} GP with a rank (at most) $m$ kernel
\begin{equation*} \label{eq_k_sor}
k_{\mathrm{SoR}}(\bm{x}_i, \bm{x}_j) = k(\bm{x}_i, \bm{X}_m) \bm{K}_{mm}^{-1} k(\bm{X}_m, \bm{x}_j).
\end{equation*}

\begin{figure*}[!htb] 
	\centering
	\includegraphics[width=0.75\textwidth]{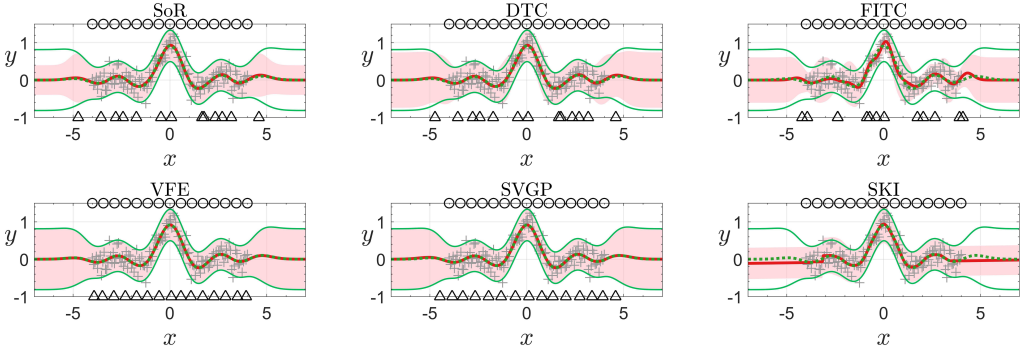}
	\caption{Illustration of sparse approximations on a 1D toy example with $y(x) = \mathrm{sinc}(x) + \epsilon$ where $\epsilon \sim \mathcal{N}(0, 0.04)$. In the panel, the $+$ symbols represent 120 training points; the top circles represent the initial locations of inducing points, whereas the bottom triangles represent the optimized locations of inducing points; the dot green curves represent the prediction mean of full GP; the green curves represent 95\% confidence interval of the full GP predictions; the red curves represent the prediction mean of sparse approximations; the shaded regions represent 95\% confidence interval of the predictions of sparse approximations. For SKI, it does not optimize over the positions of inducing points. It is found that among the three prior approximations, (i) the SoR suffers from over-confident prediction variance when leaving the training data; (ii) the FITC captures heteroscedasticity in variance; and (iii) all of them are not guaranteed to converge to the full GP, indicated by the overlapped inducing points. Differently, the VFE and its stochastic variant SVGP approximate the full GP well due to the posterior approximation. Finally, though greatly reducing the time complexity by structured inducing set, the SKI may produce discontinuous predictions.}
	\label{Fig_sparse_approxs} 
\end{figure*}

Alternatively, we could interpret the SoR from the weight-space view. It is known that the GP using a kernel with an infinite expansion of the input $\bm{x}$ in the feature space defined by dense basis functions $\{\phi_c(\bm{x})\}_{c=1}^v$ is equivalent to a Bayesian linear model in~\eqref{eq_weight_GP} with infinite weights. Hence, the relevance vector machine (RVM)~\cite{tipping2001sparse} uses only $m$ basis functions $\bm{\phi}_m(\bm{x}) = [\phi_1(\bm{x}), \cdots, \phi_m(\bm{x})]^{\mathsf{T}}$ for approximation
\begin{equation} \label{eq_weight_sparse}
p(\bm{f}|\bm{w}) = \mathcal{N}(\bm{f}| \bm{\Phi}_{nm} \bm{w}, \bm{K}_{nn} - \bm{\Phi}_{nm} \bm{\Sigma}_{mm} \bm{\Phi}_{nm}^{\mathsf{T}}), 
\end{equation}
where $\bm{\Phi}_{nm} = [\bm{\phi}_m(\bm{x}_1), \cdots, \bm{\phi}_m(\bm{x}_n)]^{\mathsf{T}}$ and $p(\bm{w}) = \mathcal{N}(\bm{w}|\bm{0}, \bm{\Sigma}_{mm})$. As a consequence, from the function-space view, the RVM is a GP with the kernel
\begin{equation*} \label{eq_k_rvm}
k_{\mathrm{RVM}}(\bm{x}_i, \bm{x}_j) = \bm{\phi}^{\mathsf{T}}(\bm{x}_i) \bm{\Sigma}_{mm} \bm{\phi}(\bm{x}_j),
\end{equation*}
which recovers $k_{\mathrm{SoR}}$ when $\bm{\Sigma}_{mm} = \bm{I}_{m}$ and $\bm{\phi}_m(\bm{x}) = \bm{L}^{\mathsf{T}} k^{\mathsf{T}}(x,\bm{X}_m)$ where $\bm{L} \bm{L}^{\mathsf{T}} = \bm{K}^{-1}_{mm}$~\cite{peng2017asynchronous}.\footnote{The choose of $\bm{\phi}_m(\bm{x})$ should produce a PSD kernel matrix such that $\bm{K}_{nn} - \bm{\Phi}_{nm} \bm{\Sigma}_{mm} \bm{\Phi}_{nm}^{\mathsf{T}} \succeq \bm{0}$. Alternatively, we can take the simple form $\bm{\phi}_m(\bm{x}) = \bm{\Lambda}_{mm} \bm{k}^{\mathsf{T}}_m(\bm{x})$ where $\bm{\Lambda}_{mm}$ is a diagonal matrix and $\bm{k}_m(\bm{x}) = k(\bm{x}, \bm{X}_m)$~\cite{tipping2001sparse}; or we take $\bm{\phi}_m(\bm{x}) = \bm{\Lambda}^{1/2} \bm{U}^{\mathsf{T}} \bm{k}^{\mathsf{T}}_m(\bm{x})$ with $\bm{\Lambda}$ and $\bm{U}$ respectively being the eigenvalue matrix and eigenvector matrix of $\bm{K}^{-1}_{mm}$~\cite{peng2015eigengp}, leading to a scaled Nystr\"{o}m approximation.} However, as depicted in Fig.~\ref{Fig_sparse_approxs}, the SoR approximation and the RVM-type models~\cite{tipping2001sparse, cressie2008fixed, peng2015eigengp} impose too restrictive assumptions to the training and test data such that they produce \textit{overconfident} prediction variances when leaving the training data.\footnote{The degenerate kernels $k_{\mathrm{SoR}}$ and $k_{\mathrm{RVM}}$ only have $m$ degrees of freedom, and suffer from the odd property of depending on inputs.} 

To reverse the uncertainty behavior of SoR, the RVM is healed through augmenting the basis functions at $\bm{x}_*$ with however higher computing cost~\cite{rasmussen2005healing}. This augmentation by including $f_*$ into $\bm{f}_m$ was also studied in~\cite{quinonero2005unifying}. Alternatively, the sparse spectrum GP (SSGP)~\cite{lazaro2010sparse} and its variational variants~\cite{gal2015improving, tan2016variational, hoang2017generalized} elegantly address this issue by reconstructing the Bayesian linear model from spectral representation (Fourier features), resulting in the \textit{stationary} kernel
\begin{equation*}
k(\bm{x}_i, \bm{x}_j) = \frac{\sigma^2_0}{m} \bm{\phi}_m^{\mathsf{T}}(\bm{x}_i) \bm{\phi}_m(\bm{x}_j) = \frac{\sigma^2_0}{m} \sum_{r=1}^m \cos \left(2\pi \bm{s}_r^{\mathsf{T}} (\bm{x}_i - \bm{x}_j) \right),
\end{equation*}
where $\bm{s}_r \in R^d$ represents the spectral frequencies. 

Another way is to impose more informative assumption to $\tilde{\bm{Q}}_{nn}$ and $\tilde{Q}_{**}$. For instance, the deterministic training conditional (DTC)~\cite{csato2002sparse, seeger2003fast} imposes the deterministic training conditional 
\begin{equation}
\label{eq_DTC_conditional}
q_{\mathrm{DTC}}(\bm{f} | \bm{f}_m) = \mathcal{N} (\bm{f}|\bm{K}_{nm} \bm{K}_{mm}^{-1} \bm{f}_m, \bm{0})
\end{equation}
but retains the exact test conditional Hence, the prediction mean is the same as that of SoR, but the prediction variance is always larger than that of SoR, and grows to the prior when leaving the inducing points, see Fig.~\ref{Fig_sparse_approxs}. Notably, due to the inconsistent conditionals in~\eqref{eq_DTC_conditional}, the DTC is not an exact GP. Besides, the DTC and SoR often perform not so well due to the restrictive prior assumption $\tilde{\bm{Q}}_{nn} = \bm{0}$.\footnote{This could be addressed by the variational variant of DTC reviewed in section~\ref{sec_post_approx}.} 

Alternatively, the fully independent training conditional (FITC)~\cite{snelson2006sparse} imposes another fully independence assumption to remove the dependency among $\{f_i\}_{i=1}^n$ such that given $\bm{V}_{nn} = \bm{K}_{nn} - \bm{Q}_{nn}$, the training conditional $q_{\mathrm{FITC}}(\bm{f} | \bm{f}_m)$
\begin{equation} \label{eq_q_FITC}
\begin{aligned}
:= \prod_{i=1}^n p(f_i | \bm{f}_m) = \mathcal{N} (\bm{f}|\bm{K}_{nm} \bm{K}_{mm}^{-1} \bm{f}_m, \mathrm{diag}[\bm{V}_{nn}]),
\end{aligned}
\end{equation}
whereas the test conditional retains exact. It is found that the variances of~\eqref{eq_q_FITC} is identical to that of $p(\bm{f} | \bm{f}_m)$ due to the correlation $\tilde{\bm{Q}}_{nn} = \mathrm{diag}[\bm{V}_{nn}]$. Hence, compared to SoR and DTC which throw away the uncertainty in~\eqref{eq_SoR_conditional} and~\eqref{eq_DTC_conditional}, FITC partially retains it, leading to a closer approximation to the prior $p(\bm{f}, f_*)$. Moreover, the fully independence assumption can be extended to $q(f_* | \bm{f}_m)$ to derive the fully independent conditional (FIC)\footnote{The predictive distributions of FITC and FIC only differ when predicting multiple test points simultaneously.}~\cite{quinonero2005unifying}, which stands as a \textit{non-degenerate} GP with the kernel
\begin{equation*}
k_{\mathrm{FIC}}(\bm{x}_i, \bm{x}_j) = k_{\mathrm{SoR}}(\bm{x}_i, \bm{x}_j) + \delta_{ij} [k(\bm{x}_i, \bm{x}_j) - k_{\mathrm{SoR}}(\bm{x}_i, \bm{x}_j)],
\end{equation*}
where $\delta_{ij}$ is the Kronecker's delta. Note that $k_{\mathrm{FIC}}$ has a constant prior variance but is not stationary. Alternatively, the approximation in~\eqref{eq_q_FITC} can be derived from minimizing the Kullback-Leibler (KL) divergence $\mathrm{KL} (p(\bm{f}, \bm{f}_m) || q(\bm{f}_m) \prod_{i=1}^n q(f_i|\bm{f}_m))$~\cite{snelson2008flexible}, which quantifies the similarity between the exact and approximated joint prior.

Particularly, the FITC produces prediction mean and variance at $\bm{x}_*$ as $\mu(\bm{x}_*) = \bm{k}_{*m} \bm{\Psi} \bm{K}_{mn} \bm{\Lambda}^{-1} \bm{y}$ and $\sigma^2(\bm{x}_*) = k_{**} - Q_{**} + \bm{k}_{*m} \bm{\Psi} \bm{k}_{m*}$， where $\bm{\Psi}^{-1} = \bm{K}_{mm} + \bm{K}_{mn} \bm{\Xi}^{-1} \bm{K}_{nm}$ and $\bm{\Xi} = \mathrm{diag}[\bm{V}_{nn}] + \sigma^2_{\epsilon} \bm{I}_n$. It is found that  the diagonal correlation $\mathrm{diag}[\bm{V}_{nn}]$ represents the posterior variances of $\bm{f}$ given $\bm{f}_m$. Hence, these varying variances, which are zeros exactly at $\bm{X}_m$, enable FITC to capture the noise heteroscedasticity, see Fig.~\ref{Fig_sparse_approxs}, at the cost of (i) producing an invalidate estimation (nearly zero) of the noise variance $\sigma^2_{\epsilon}$, and (ii) sacrificing the accuracy of prediction mean~\cite{bauer2016understanding}. 

To improve FITC, the partially independent training conditional (PITC)~\cite{quinonero2005unifying} has the training conditional $q_{\mathrm{PITC}}(\bm{f} | \bm{f}_m)$
\begin{equation} \label{eq_q_PITC}
\begin{aligned}
:= \prod_{i=1}^M p(\bm{f}_i | \bm{f}_m) = \mathcal{N} (\bm{f}|\bm{K}_{nm} \bm{K}_{mm}^{-1} \bm{f}_m, \mathrm{blkdiag}[\bm{V}_{nn}]).
\end{aligned}
\end{equation}
This equates to partitioning the training data $\mathcal{D}$ into $M$ independent subsets (blocks) $\{ \mathcal{D}_i \}_{i=1}^M$, and taking into account the joint distribution of $\bm{f}_i$ in each subset. But it is argued that though being a closer approximation to $p(\bm{f} | \bm{f}_m)$, the blocking $q_{\mathrm{PITC}}(\bm{f} | \bm{f}_m)$ brings little improvements over FITC~\cite{snelson2007local}. This issue can be addressed by the extended partially independent conditional (PIC)~\cite{snelson2007local} discussed in section~\ref{sec_model_capability}.

So far, we have reviewed state-of-the-art prior approximations including SoR, DTC, FI(T)C and PITC. Regarding their implementations, the choose of inducing points is crucial. Alternatively, similar to SoD, we could use clustering techniques to select a finite set of space-filling inducing points from $\mathcal{D}$, or we employ some querying criteria~\cite{csato2001sparse, smola2001sparse, seeger2003fast, keerthi2006matching, schreiter2016efficient} to sequentially choose informative inducing points. More flexibly, inducing points are regarded as parameters to be optimized together with other hyperparameters~\cite{snelson2006sparse}, which additionally introduces $m \times d$ parameters and turns the inference into a high-dimensional optimization task. Besides, with increasing $m$, the benefits brought by the optimization over the simple selection from training data vanish. Interestingly, a recent work~\cite{pourhabib2014bayesian} shows the first attempt to simultaneously determine the number and locations of inducing points in the Bayesian framework by placing a prior on $\bm{X}_m$.

Finally, the heteroscedasticity of FITC raises another finding that this approximation attempts to achieve a desirable predictive accuracy at low computing cost, rather than faithfully recovering the standard GP with increasing $m$. Indeed, the prior approximations recover the full GP when $\bm{X}_m = \bm{X}$. But this configuration is not the global optimum when maximizing $\log q(\bm{y})$, which makes them philosophically troubling. Besides, learning inducing points via the optimization of~\eqref{eq_log_q(y)} may produce poor predictions~\cite{titsias2009variational}. These issues will be addressed by the posterior approximations reviewed below.

\vspace{1ex}
\subsubsection{\textbf{Posterior approximations}}
\label{sec_post_approx}
Different from prior approximations, posterior approximations~\cite{titsias2009variational, hensman2013gaussian} retain exact prior but perform approximate inference. The most well-known posterior approximation is the elegant variational free energy (VFE)~\cite{titsias2009variational} proposed by Titsias in 2009 by using variational inference (VI)~\cite{blei2017variational}. Instead of modifying the prior $p(\bm{f}, f_*)$, VFE directly approximates the posterior $p(\bm{f}, \bm{f}_m | \bm{y})$, the learning of which is a central task in statistical models, by introducing a variational distribution $q(\bm{f}, \bm{f}_m | \bm{y})$. Then, we have their KL divergence $\mathrm{KL}(q(\bm{f}, \bm{f}_m | \bm{y}) || p(\bm{f}, \bm{f}_m | \bm{y}))$
\begin{equation}
\begin{aligned}
:= & \log p(\bm{y}) - \left\langle   \log \frac{p(\bm{y}, \bm{f}, \bm{f}_m)}{q(\bm{f}, \bm{f}_m | \bm{y})} \right\rangle_{q(\bm{f}, \bm{f}_m | \bm{y})} = \log p(\bm{y}) - F_q,
\end{aligned}
\end{equation}
where $\left\langle . \right\rangle_{q(.)} $ represents the expectation over the distribution $q(.)$.\footnote{Matthews et al.~\cite{matthews2016sparse} further extended the procedure to infinite index sets using the KL divergence between stochastic processes such that the posterior is approximated over the entire process $f$.} It is found that minimizing the rigorously defined $\mathrm{KL}(q || p) \ge 0$ is equivalent to maximizing $F_q$, since $\log p(\bm{y})$ is constant for $q(\bm{f}, \bm{f}_m | \bm{y})$. Thus, $F_q$ is called evidence lower bound (ELBO) or variational free energy, which permits us to jointly optimize the variational parameters\footnote{Notably, the inducing positions are regarded as the variational parameters in $q(\bm{f}_m | \bm{y})$ rather than the model parameters.} and hyperparameters. It is observed that maximizing $F_q$ w.r.t. the hyperparameters directly improves $F_q$; while maximizing $F_q$ w.r.t. the variational parameters implicitly drives the approximation to match both the posterior $p(\bm{f}, \bm{f}_m | \bm{y})$ and the evidence $p(\bm{y})$.

To derive a tighter bound, the calculus of variations finds the optimal variational distribution $q^*(\bm{f}_m | \bm{y})$ to remove the dependency of $F_q$ on $q(\bm{f}_m | \bm{y})$ by taking the relevant derivative to zero, leading to the ``collapsed'' bound
\begin{equation} \label{eq_F_q}
\begin{aligned}
F_{\mathrm{VFE}} =& \log q_{\mathrm{DTC}} (\bm{y}) - \frac{1}{2 \sigma^2_{\epsilon}} \mathrm{tr} [\bm{V}_{nn}] \ge F_q.
\end{aligned}
\end{equation} 
Note that $F_{\mathrm{VFE}}$ differs with $\log q_{\mathrm{DTC}}$ only by a trace term, which however substantially improves the inference quality. In order to maximize $F_{\mathrm{VFE}}$, we should decrease the trace $\mathrm{tr} [\bm{V}_{nn}] \ge 0$, which represents the total variance of predicting the latent variables $\bm{f}$ given $\bm{f}_m$. Particularly, $\mathrm{tr} [\bm{V}_{nn}] = 0$ means $\bm{f}_m = \bm{f}$ and we recover the full GP. Hence, the trace term (i) is a regularizer that guards against over-fitting; (ii) seeks to deliver a good inducing set; and (iii) always improves $F_q$ with increasing $m$, see the theoretical analysis in~\cite{bauer2016understanding, matthews2016sparse}. The third property implies that given enough resources the VFE will recover the full GP, see Fig.~\ref{Fig_sparse_approxs}. In contrast, without this trace term, the DTC often risks over-fitting~\cite{titsias2009model}. 

Regarding the improvements of VFE, it was extended to continuous and discrete inputs through an efficient QR factorization-based optimization over both inducing points and hyperparameters~\cite{cao2015efficient}. The estimation of inducing points has also been improved in an augmented feature space~\cite{zhe2017regularized}, which is similar to the inter-domain strategy~\cite{lazaro2009inter}. The authors argued that the similarity of inducing points measured in the Euclidean space is inconsistent to that measured by the GP kernel function. Hence, they assigned a mixture prior on $\bm{X}$ in the latent feature space, and derived a regularized bound for choosing good inducing points in the kernel space. Besides, Matthews et al.~\cite{matthews2016sparse, de2016scalable} bridged the gap between the variational inducing-points framework and the more general KL divergence between stochastic processes. Using this new interpretation, Bui et al.~\cite{bui2017unifying} approximated the general, infinite joint prior $p(\bm{f}_m, f_{\ne \bm{f}_m}, \bm{y}) = p(\bm{f}_m, f_{\ne \bm{f}_m} | \bm{y}) p(\bm{y})$ which comprises two inferential objects of interest: posterior distribution and model evidence. Minimizing their KL divergence thus encourages \textit{direct} approximation to both posterior and evidence. Hence, the FITC and VFE are interpreted jointly as
\begin{equation}
\log q_{\mathrm{PEP}}(\bm{y}) = \log q(\bm{y}) - \frac{1 - \alpha}{2 \alpha} \mathrm{tr} \left[ \log \left(\bm{I}_n + \frac{\alpha}{\sigma^2_{\epsilon}} \bm{V}_{nn} \right) \right],
\end{equation}
where $\log q(\bm{y})$ takes the form~\eqref{eq_log_q(y)} with $\tilde{\bm{Q}}_{nn} = \alpha \mathrm{diag}[\bm{V}_{nn}]$. By varying $\alpha \in (0,1]$,
we recover FITC when $\alpha = 1$ and VFE when $\alpha \to 0$. Besides, a hybrid approximation using a moderate $\alpha$, e.g., $\alpha = 0.5$, often produces better predictions.

To further improve the scalability of VFE, Hensman et al.~\cite{hensman2013gaussian} retained the variational distribution $q(\bm{f}_m|\bm{y}) = \mathcal{N}(\bm{f}_m| \bm{m}, \bm{S})$ in $F_q$ to obtain a relaxed bound
\begin{equation} \label{eq_F_Q}
F_q = \left\langle \log p(\bm{y}|\bm{f}) \right\rangle_{p(\bm{f}|\bm{f}_m) q(\bm{f}_m | \bm{y})} - \mathrm{KL}(q(\bm{f}_m|\bm{y}) || p(\bm{f}_m)).
\end{equation}
The first term in the right-hand side of $F_q$ is the sum of $n$ terms due to the \textit{iid} observation noises, i.e., $p(\bm{y} | \bm{f}) = \prod_{i=1}^n p(y_i | f_i)$. Hence, the stochastic gradient descent (SGD)~\cite{kingma2014adam}, which encourages large-scale learning, could be employed to obtain an unbiased estimation of $F_q$ using a mini-batch $\{ \bm{X}_b, \bm{y}_b \}$ as
\begin{equation} \label{eq_F_Q_minibatch}
\begin{aligned}
F_q \approx & \frac{n}{|\bm{y}_b|} \sum_{y_i \in \bm{y}_b} \int q(\bm{f}_m|\bm{y}) p(f_i|\bm{f}_m) \log p(y_i|f_i) df_i d\bm{f}_m
 \\
&- \mathrm{KL}(q(\bm{f}_m|\bm{y}) || p(\bm{f}_m)).
\end{aligned}
\end{equation}
Due to the difficulty of optimizing variational parameters $\bm{m}$ and $\bm{S}$ in the Euclidean space, one can employ the Stochastic Variational Inference (SVI)~\cite{hoffman2013stochastic} using natural gradients,\footnote{The superiority of natural gradients over ordinal gradients for regression has been verified in~\cite{salimbeni2018natural}.} resulting in a remarkable complexity of $\mathcal{O}(m^3)$ when $|\bm{y}_b|=1$, and more interestingly, the \textit{online} or \textit{anytime} learning fashion. 

Therefore, a crucial property of the stochastic variational GP (SVGP) is that it trains a sparse GP at \textit{any time} with a small subset of the training data in each iteration~\cite{hoang2015unifying}. Another interesting property is that taking a unit step in the natural gradient direction equals to performing an update in the Variational Bayes Expectation Maximization (VB-EM) framework~\cite{hensman2012fast}. Though showing high scalability and desirable approximation, the SVGP has some drawbacks: (i) the bound $F_q$ is less tight than $F_{\mathrm{VFE}}$ because $q(\bm{f}_m|\bm{y})$ is not optimally eliminated; (ii) it optimizes over $q(\bm{f}_m|\bm{y})$ with a huge number of variational parameters, thus requiring much time to complete one epoch of training; and (iii) the introduction of SVI brings the empirical requirement of carefully turning the parameters of SGD.

Inspired by the idea of Hensman, Peng et al.~\cite{peng2017asynchronous} derived the similar factorized variational bound for GPs by taking the weight-space augmentation in~\eqref{eq_weight_sparse}. The weight-space view (i) allows using flexible basis functions to incorporate various low-rank structures; and (ii) provides a composite non-convex bound enabling the speedup using an asynchronous proximal gradient-based algorithm~\cite{li2013distributed}. By deploying the variational model in a distributed machine learning platform PARAMETERSERVER~\cite{li2014communication}, the authors have first scaled GP up to \textit{billions} of data points. Similarly, Cheng and Boots~\cite{cheng2017variational} also derived a stochastic variational framework from the weight-space view with the difference being that the mean and variance of $p(\bm{f} | \bm{w})$ respectively use the \textit{decoupled} basis function sets $\bm{\phi}_a$ and $\bm{\phi}_b$, leading to more flexible inference. Besides, a recent interesting work~\cite{hoang2015unifying} presents a novel unifying, \textit{anytime} variational framework akin to Hensman's for accommodating existing sparse approximations, e.g., SoR, DTC, FIT(C) and PIT(C), such that they can be trained via the efficient SGD which achieves asymptotic convergence to the predictive distribution of the chosen sparse model. The key of this work is to conduct a \textit{reverse} variational inference wherein ``reverse'' means we can find a prior $p(\bm{f}_m) = \mathcal{N}(\bm{f}_m|\bm{}\nu, \bm{\Lambda})$ (not the conventional GP prior) such that the variational distribution $q^*(\bm{f}_m | \bm{y}) = p(\bm{f}_m | \bm{y})$ for FI(T)C and PI(T)C is the maximum of the variational lower bound.\footnote{For VFE and its stochastic variant SVGP, normally, we pre-define the prior $p(\bm{f}_m) = \mathcal{N}(\bm{f}_m|\bm{0}, \bm{K}_{mm})$, and then find an optimal $q^*(\bm{f}_m | \bm{y})$ to maximize the variational lower bound.} Finally, the scalability of Hensman's model can be further reduced to nearly $\mathcal{O}(m)$ by introducing Kronecker structures for inducing points and the variance of $q(\bm{f}_m|\bm{y})$~\cite{nickson2015blitzkriging, izmailov2017scalable}.

Titsias and Hensman's models have been further improved by using, e.g., (i) Bayesian treatment of hyperparameters~\cite{aueb2013variational, hensman2015mcmc, yu2017stochastic} rather than traditional point estimation which risks over-fitting when the number of hyperparameters is small; and (ii) non-Gaussian likelihoods~\cite{hensman2015mcmc, sheth2015sparse, hensman2016variational}.

\vspace{1ex}
\subsubsection{\textbf{Structured sparse approximations}}
A direct speedup to solve $(\bm{K}_{nn}^{\epsilon})^{-1} \bm{y}$ in standard GP can be achieved through fast matrix-vector multiplication (MVM)~\cite{shen2006fast, morariu2009automatic}, which iteratively solves the linear system using conjugate gradients (CG) with $s$ ($s \ll n$) iterations,\footnote{The solution to $\bm{A} \bm{x} = \bm{b}$ is the unique minimum of the quadratic  function $0.5 \bm{x}^{\mathsf{T}} \bm{A} \bm{x} - \bm{x}^{\mathsf{T}} \bm{b}$.} resulting in a time complexity of $\mathcal{O}(sn^2)$. It was argued by~\cite{chalupka2013framework} that the original MVM has some open questions, e.g., the determination of $s$, the lack of meaningful speedups, and the badly conditioned kernel matrix. Alternatively, the pre-conditioned CG (PCG)~\cite{cutajar2016preconditioning} employs a pre-conditioning matrix through for example the Nystr\"{o}m approximation to improve the conditioning of kernel matrix and accelerate the CG convergence. 

More interestingly, when the kernel matrix $\bm{K}_{nn}$ itself has some algebraic structure, the MVM provides massive scalability. For example, the Kronecker methods~\cite{saatcci2012scalable, gilboa2015scaling} exploit the multi-variate grid inputs $\bm{x} \in \Omega_1 \times \cdots \times \Omega_d$ and the tensor product kernel with the form $k(\bm{x}_i, \bm{x}_j) = \prod_{t=1}^d k(\bm{x}_i^t, \bm{x}_j^t)$.\footnote{Popular kernels for example the SE kernel~\eqref{eq_SE} fulfill the product structure.} Then, the kernel matrix decomposes to a Kronecker product $\bm{K}_{nn} = \bm{K}_1 \otimes \cdots \otimes \bm{K}_d$, which eases the eigen-decomposition with a greatly reduced time complexity of $\mathcal{O}(d\overline{n}^{d+1})$ where $\overline{n} = \sqrt[d]{n}$ for $d > 1$.\footnote{$\bm{K}_i$ ($1 \le i \le d$) is an $\overline{n} \times \overline{n}$ matrix when the number of points along each dimension is the same. Besides, for the detailed introduction of GP with multiplicative kernels, please refer to section 2.2 of~\cite{gilboa2015scaling}.} Another one is the Toeplitz methods~\cite{cunningham2008fast}---complementary to the Kronecker methods---that exploit the kernel matrix built from regularly spaced one dimensional points, resulting in the time complexity of $\mathcal{O}(d \overline{n}^d \log \overline{n})$. The severe limitation of the Kronecker and Toeplitz methods is that they require grid inputs, preventing them from being applied to the general arbitrary data points.\footnote{This limitation was relaxed in the \textit{partial} \textit{grid} and variable noise scenario by introducing virtual observations and inputs~\cite{wilson2014fast, gilboa2015scaling}.}

To handle arbitrary data while retaining the efficient Kronecker structure, the structured kernel interpolation (SKI)~\cite{wilson2015kernel} imposes the grid constraint on the inducing points. Hence, the matrix $\bm{K}_{mm}$ admits the Kronecker structure for $d > 1$ and the Toeplitz structure for $d=1$, whereas the cross kernel matrix $\bm{K}_{nm}$ is approximated for example by a local linear interpolation using adjacent grid inducing points as
\begin{equation} \label{eq_SKI}
k (\bm{x}_i, \bm{u}_j) \approx w_i k(\bm{u}_a, \bm{u}_j) + (1-w_i) k(\bm{u}_b, \bm{u}_j),
\end{equation}
where $\bm{u}_a$ and $\bm{u}_b$ are two inducing points most closely bound $\bm{x}_i$, and $w_i$ is the interpolation weight. Inserting the approximation~\eqref{eq_SKI} back into $\bm{Q}_{nn}$, we have
\begin{equation} \label{eq_SKI_Qnn}
\bm{Q}_{nn} \approx \bm{W}_{nm} \bm{K}_{mm}^{-1} \bm{W}_{nm}^{\mathsf{T}},
\end{equation}
where the weight matrix $\bm{W}$ is extremely sparse since it only has two non-zero entires per row for local linear interpolation, leading to an impressive time complexity of $\mathcal{O}(n + d \overline{m}^{d+1})$ with $\overline{m} = \sqrt[d]{m}$ for solving $(\bm{K}_{nn}^{\epsilon})^{-1} \bm{y}$. Also, the sparse $\bm{W}$ incurs the prediction mean with \textit{constant-time} complexity $\mathcal{O}(1)$ and the prediction variance with complexity $\mathcal{O}(m)$ after pre-computing. Furthermore, Pleiss et al.~\cite{pleiss2018constant} derived a \textit{constant-time} prediction variance using Lanczos approximation, which admits $s$ iterations of MVM for calculation.

The original SKI has two main drawbacks. First, the number $m$ of \textit{grid} inducing points grows \textit{exponentially} with dimensionality $d$, making it impractical for $d > 5$. To address this issue, one could use dimensionality reduction or manifold learning to map the inducing points into a $p$-dimensional ($p \ll d$) latent space~\cite{wilson2015thoughts}; or more interestingly, one can use the hierarchical structure of neural networks to extract the latent low-dimensional feature space~\cite{wilson2016deep, wilson2016stochastic}. Furthermore, continual efforts~\cite{evans2017scalable, izmailov2017scalable, gardner2018product} have been made to directly reduce the time complexity to be \textit{linear} with $d$ by exploiting the row-partitioned Khatri-Rao structure of $\bm{K}_{nm}$, or imposing tensor train decomposition and Kronecker product to the mean and variance of $q(\bm{f}_m|\bm{y})$ in Hensman's variational framework. The linear complexity with $d$ permits the use of numerous inducing points, e.g., $m = 10^d$.

Second, the SKI may produce discontinuous predictions due to the local weight interpolation, and provide overconfident prediction variance when leaving the training data due to the restrictive SoR framework, see Fig.~\ref{Fig_sparse_approxs}. To smooth the predictions, Evans and Nair~\cite{evans2017scalable} exploited the row-partitioned Khatri-Rao structure of $\bm{K}_{nm}$ rather than using local weight interpolation. To have sensible uncertainty, a diagonal correlation akin to that of FITC has been considered~\cite{evans2017scalable, dong2017scalable}.

Finally, note that the permit of many inducing points is expected to improve the model capability. But due to the \textit{grid} constraint, the structured sparse approximations (i) use fixed inducing points, (ii) resort to dimensionality reduction for tackling high-dimensional tasks, and (iii) place the vast majority of inducing points on the domain boundary with increasing $d$, which in turn may degenerate the model capability.

\section{Local approximations} 
\label{sec_local_approx}
Inspired by D\&C, local approximations use localized experts to improve the scalability of GP. Besides, compared to global approximations, the local nature enables capturing \textit{non-stationary} features. In what follows, we comprehensively review the \textit{naive-local-experts} which directly employs the pure local experts for prediction, and the \textit{mixture-of-experts} and \textit{product-of-experts} which inherit the advantages of naive-local-experts but boost the predictions through model averaging.

\subsection{Naive-local-experts}
It is known that a pair of points far away from each other has a low correlation. Hence, localized experts trained on subsets of $\mathcal{D}$ is expected to produce sensible predictions with low computational complexity. Particularly, the simple naive-local-experts (NLE)~\cite{kim2005analyzing,datta2016nearest} lets the local expert $\mathcal{M}_i$ completely responsible for the subregion $\Omega_i$ defined by $\bm{X}_i$. Mathematically, we predict at $\bm{x}_* \in \Omega_i$ as $p(y_*|\mathcal{D}, \bm{x}_*) \approx p_i(y_*|\mathcal{D}_i, \bm{x}_*)$.

According to the partition of $\mathcal{D}$, we classify NLE into two main categories: (i) \textit{inductive} NLE, which first partitions the input space and trains all the experts, and then chooses an appropriate one for predicting at $\bm{x}_*$; and (ii) \textit{transductive} NLE, which particularly chooses a neighborhood subset $\mathcal{D}_*$ around $\bm{x}_*$, and trains the relevant expert $\mathcal{M}_*$ for predicting at $\bm{x}_*$. 

Inductive NLE employs a \textit{static} partition of the whole data using clustering techniques, e.g., Voronoi tessellations~\cite{kim2005analyzing} and trees~\cite{vasudevan2009gaussian, pratola2014parallel}, and trains independent local GP experts, resulting in $\mathcal{O}(nm_0^2)$ where $m_0 = n/M$ is the training size for each expert. The partition and the experts are usually learned separately; or they can be learned jointly with Bayesian treatment~\cite{denison2002bayesian}. In contrast, transductive NLE, e.g., the nearest-neighbors (NeNe)~\cite{urtasun2008sparse} which could induce a valid stochastic process~\cite{datta2016hierarchical, datta2016nearest}, employs a \textit{dynamic} partition to choose $m_0$ neighbor points around $\bm{x}_*$, resulting in $\mathcal{O}(n_t m_0^3)$ complexity that relies on the test size $n_t$. A key issue in transductive NLE is the definition of the neighborhood set $\mathcal{D}_*$ around $\bm{x}_*$. The simplest way is using geometric closeness criteria\footnote{The selected points should be close to $\bm{x}_*$; meanwhile, they should distribute uniformly to avoid conveying redundant information.} for selection, which however are not optimal without considering the spatial correlation~\cite{emery2009kriging}. Hence, some GP-based active learning methods have been employed to sequentially update the neighborhood set~\cite{gramacy2009adaptive, gramacy2015local, gramacy2016lagp, gramacy2016speeding}.

\begin{figure*}[!htb] 
	\centering
	\includegraphics[width=0.7\textwidth]{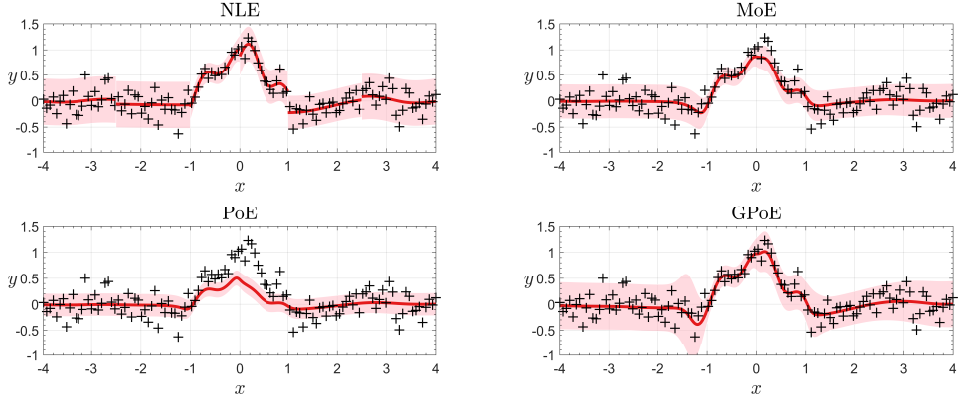}
	\caption{Illustration of local approximations using six individual experts on the toy example. Note that for MoE, we did not jointly learn the experts and gating functions. For simplicity, we use the individual experts and the differential entropy as $\beta_i$ in the softmax gating function. It is found that the NLE suffers from discontinuity and poor generalization. The PoE produces poor prediction mean and overconfident prediction variance  due to the inability of suppressing poor experts. To alleviate this issue, we could either use gating functions, like the MoE, to provide desirable predictions; or use input-dependent weights, like GPoE, to boost the predictions. }
	\label{Fig_local_approxs} 
\end{figure*}

Whilst enjoying the capability of capturing non-stationary features due to the localized structure, the NLE (i) produces discontinuous predictions on the boundaries of subregions and (ii) suffers from poor generalization capability since it misses the long-term spatial correlations, as depicted in Fig.~\ref{Fig_local_approxs}. To address the discontinuity issue, the patched GPs~\cite{park2016efficient, park2017patchwork} impose continuity conditions such that two adjacent local GPs are patched to share the nearly identical predictions on the boundary. But the patched GPs suffer from inconsistent and even negative prediction variance, and are only available in low dimensional space~\cite{pourhabib2014bayesian, park2017patchwork}. More popularly, the model averaging strategy, which is accomplished by the mixture/product of local GP experts elaborated below, well smooths the predictions from multiple experts. To address the generalization issue, it is possible to (i) share the hyperparameters across experts, like~\cite{deisenroth2015distributed}; or (ii) combine local approximations with global approximations, which will be reviewed in section~\ref{sec_model_capability}.

\subsection{Mixture-of-experts}
\begin{figure}[!htb] 
	\centering
	\includegraphics[width=0.22\textwidth]{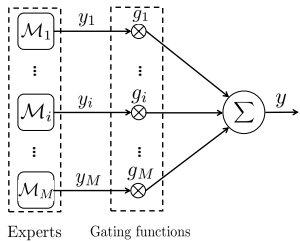}
	\caption{Illustration of mixture-of-experts.}
	\label{Fig_MoE} 
\end{figure}

The mixture-of-experts (MoE) devotes to combining the \textit{local} and \textit{diverse} experts owning individual hyperparameters for improving the overall accuracy and reliability~\cite{yuksel2012twenty, masoudnia2014mixture}.\footnote{This topic has been studied in a broad regime, called \textit{ensemble learning}~\cite{mendes2012ensemble}, for aggregating various learning models to boost predictions.} As shown in Fig.~\ref{Fig_MoE}, MoE generally expresses the combination as a Gaussian mixture model (GMM)~\cite{jacobs1991adaptive}
\begin{equation} \label{eq_MoE}
p(y|\bm{x}) = \sum_{i=1}^M g_i(\bm{x}) p_i(y|\bm{x}),
\end{equation}
where $g_i(\bm{x})$ is the gating function, which usually takes a parametric form like the softmax or probit function~\cite{jacobs1991adaptive, geweke2007smoothly}, and can be thought as the probability $p(z = i) = \pi_i$ that the expert indicator $z$ is $i$, i.e., $\bm{x}$ is assigned to expert $\mathcal{M}_i$; $p_i(y|\bm{x})$ comes from $\mathcal{M}_i$ and is responsible for component $i$ of the mixture. In~\eqref{eq_MoE}, the gates $\{g_i\}_{i=1}^M$ manage the mixture through \textit{probabilistic} partition of the input space for defining the subregions where the individual experts responsible for. The experts can be a variety of machine learning models, e.g., linear model and support vector machines~\cite{ xu1995alternative, lima2007hybridizing}. 

The training of MoE usually assumes that the data is \textit{iid} such that we maximize the factorized log likelihood $\sum_{t=1}^n \log p(y_t|\bm{x}_t)$ to learn the gating functions and the experts simultaneously by the gradient-based optimizers~\cite{chen1999improved} and more popularly, the EM algorithm~\cite{jacobs1991adaptive, jordan1994hierarchical, xu1995alternative, ng2007extension}. The joint learning permits (i) probabilistic (soft) partition of the input space via both the data and the experts themselves, and (ii) diverse experts specified for different but \textit{overlapped} subregions. Finally, the predictive distribution at $\bm{x}_*$ is
\begin{equation}
p(y_*|\mathcal{D},\bm{x}_*) = \sum_{i=1}^M g_i(\bm{x}_*|\mathcal{D}) p_i(y_*|\mathcal{D}, \bm{x}_*),
\end{equation}
where $g_i(\bm{x}_*|\mathcal{D})$ can be regarded as the posterior probability $p(z_*=i|\mathcal{D})$, called responsibility.

To advance the MoE, (i) the single-layer model in Fig.~\ref{Fig_MoE} is extended to a tree-structured hierarchical architecture~\cite{jordan1994hierarchical}; (ii) the Bayesian approach is employed instead of the maximum likelihood to get rid of over-fitting and noise-level underestimate~\cite{bishop2002bayesian}; (iii) the \textit{t}-distribution is considered to handle the outliers~\cite{chamroukhi2017skew}; and finally (iv) instead of following the conditional mixture~\eqref{eq_MoE}, the input distribution $p(\bm{x})$ is considered to form the joint likelihood $p(y, \bm{x})$ for better assignment of experts~\cite{xu1995alternative}. 

Next, we review the mixture of GP experts for big data. It is observed that (i) the original MoE is designed for capturing multi-modal (non-stationary) features, and the individual \textit{global} experts are responsible for all the data points, leading to high complexity; (ii) the \textit{iid} data assumption does not hold for GP experts which model the data dependencies through joint distribution; and (iii) the parametric gating function $g_i$ is not favored in the Bayesian non-parametric framework. In 2001, Tresp~\cite{tresp2001mixtures} first introduced the mixture of GP experts, which employs $3M$ GP experts to respectively capture the mean, the noise variance, and the gate parameters with nearly $\mathcal{O}(3Mn^3)$ complexity, which is unattractive for big data. 

The mixture of GP experts for big data should address two issues: (i) how to reduce the computational complexity (model complexity), and (ii) how to determine the number of experts (model selection).

To address the model complexity issue, there are three core threads. The first is the localization of experts. For instance, the infinite mixture of GP experts (iMGPE)~\cite{rasmussen2002infinite} uses a localized likelihood to get rid of the \textit{iid} assumption as
\begin{equation}  \label{eq_MoE_p(y)}
p(\bm{y}|\bm{X}) = \sum_{\bm{z} \in \mathcal{Z}} p(\bm{z}|\bm{X}) \prod_{i} p(\bm{y}_i|\bm{z}, \bm{X}_i).
\end{equation}
Given an instance of the expert indicators $\bm{z} = [z_1, \cdots, z_n]^{\mathsf{T}}$, the likelihood factorizes over local experts, resulting in $\mathcal{O}(nm_0^2)$ when each expert has the same training size $m_0$. Similar to~\cite{xu1995alternative}, the iMGPE model was further improved by employing the joint distribution $p(\bm{y}, \bm{X})$ rather than the conditional $p(\bm{y}|\bm{X})$ as~\cite{meeds2006alternative}
\begin{equation} \label{eq_MoE_p(y,X)}
p(\bm{y}, \bm{X}) = \sum_{\bm{z} \in \mathcal{Z}} p(\bm{z}) \prod_{i} p(\bm{y}_i|\bm{z}, \bm{X}_i) p(\bm{X}_i|\bm{z}).
\end{equation}
The fully generative model is capable of handling partially specified data and providing inverse functional mappings. But the inference over~\eqref{eq_MoE_p(y)} and~\eqref{eq_MoE_p(y,X)} should resort to the expensive Markov Chain Monte Carlo (MCMC) sampling. Alternatively, the localization can be achieved by the \textit{hard-cut} EM algorithm using a truncation representation, wherein the E-step assigns the data to experts through maximum a posteriori (MAP) of the expert indicators $\bm{z}$ or a threshold value~\cite{yang2011efficient, nguyen2014fast, chen2014precise, nguyen2016variational}. Thereafter, the M-step only operates on small subsets. 

The second thread is combining global experts with the sparse approximations reviewed in section~\ref{sec_sparse_approximations} under the variational EM framework. The dependency among outputs is broken to make variational inference feasible by (i) interpreting GP as the finite Bayesian linear model in~\eqref{eq_weight_sparse}~\cite{yuan2009variational, sun2011variational}, or (ii) using the FITC experts that factorize over $\bm{f}$ given the inducing set $\bm{f}_m$~\cite{nguyen2014fast, nguyen2016variational}. With $m$ inducing points for each expert, the complexity is $\mathcal{O}(nm^2M)$, which can be further reduced to $\mathcal{O}(nm^2)$ with the hard-cut EM~\cite{nguyen2014fast, nguyen2016variational}.

Note that the first two threads assign the data dynamically according to the data property and the experts' performance. Hence, they are denoted as mixture of implicitly localized experts (MILE)~\cite{masoudnia2014mixture}. The implicit partition determines the optimal allocation of experts, thus enabling capturing the interaction among experts. This advantage encourages the application on data association~\cite{lazaro2012overlapping, ross2013nonparametric}. The main drawback of MILE however is that in the competitive learning process, some experts may be eliminated due to the zero-coefficient problem caused by unreasonable initial parameters~\cite{hansen2000combining}. 

To relieve the problem of MILE, the third thread is to pre-partition the input space by clustering techniques and assign points to the experts before model training~\cite{nguyen2009model, liu2016kinect}. The mixture of explicitly localized experts (MELE)~\cite{masoudnia2014mixture} (i) reduces the model complexity as well, and (ii) explicitly determines the architecture of MoE and poses distinct local experts. In the meantime, the drawback of MELE is that the clustering-based partition misses the information from data labels and experts such that it cannot capture the interaction among experts.

Finally, to address the model selection issue, the Akaike information criterion~\cite{huang2014estimating} and the synchronously balancing criterion~\cite{zhao2015effective} have been employed to choose over a set of candidate $M$ values. More elegantly, the input-dependent Dirichlet process (DP)~\cite{rasmussen2002infinite}, the Polya urn distribution~\cite{meeds2006alternative} or the more general Pitman-Yor process~\cite{chatzis2012nonparametric} is introduced over the expert indicators $\bm{z}$ to act as gating functions, which in turn automatically infer the number of experts from data. The complex prior and the infinite $M$ however raise the difficulty in inference~\cite{sun2011variational}. Therefore, for simplicity, a stick-breaking representation of DP is usually used~\cite{nguyen2016variational, sun2011variational}.

\subsection{Product-of-experts}
Different from the MoE which employs a weighted sum of several probability distributions (experts) via an ``\textit{or}'' operation, the product-of-experts (PoE)~\cite{hinton2002training} multiplies these probability distributions, which sidesteps the weight assignment in MoE and is similar to an ``\textit{and}'' operation, as
\begin{equation} \label{eq_PoE}
p(y|\bm{x}) = \frac{1}{Z} \prod_{i=1}^M p_i(y|\bm{x}),
\end{equation}
where $Z$ is a normalizer, which however makes the inference intractable when maximizing the likelihood $\sum_{i=1}^n \log p(y_i|\bm{x}_i)$~\cite{hinton2006training}.

Fortunately, the GP experts sidestep this issue since $p_i(y|\bm{x})$ in~\eqref{eq_PoE} is a Gaussian distribution. Hence, the product of multiple Gaussians is still a Gaussian distribution, resulting in a factorized marginal likelihood of PoE over GP experts
\begin{equation} \label{eq_lik_PoE}
p(\bm{y}|\bm{X}) = \prod_{i=1}^M p(\bm{y}_i|\bm{X}_i),
\end{equation}
where $p_i(\bm{y}_i|\bm{X}_i) \sim \mathcal{N}(\bm{y}_i|\mathbf{0}, \bm{K}_i + \sigma^2_{\epsilon, i} \bm{I}_{n_i})$ with $\bm{K}_i = k(\bm{X}_i, \bm{X}_i) \in R^{n_i \times n_i}$ and $n_i$ being the training size of expert $\mathcal{M}_i$. This factorization degenerates the full kernel matrix $\bm{K}_{nn}$ into a diagonal block matrix $\mathrm{diag}[\bm{K}_1, \cdots, \bm{K}_M]$, leading to $\bm{K}^{-1}_{nn} \approx \mathrm{diag}[\bm{K}_1^{-1}, \cdots, \bm{K}_M^{-1}]$. Hence, the complexity is substantially reduced to $\mathcal{O}(nm_0^2)$ given $n_i = m_0$. 

It is observed that the PoE likelihood~\eqref{eq_lik_PoE} is a special case of the MoE likelihood~\eqref{eq_MoE_p(y)}: the MoE likelihood averages the PoE likehood over possible configurations of the expert indicators $\bm{z}$. Consequently, the joint learning of gating functions and experts makes MoE achieve optimal allocation of experts such that it may outperform PoE~\cite{cao2014generalized}. Generally, due to the weighted sum form~\eqref{eq_MoE}, the MoE will never be sharper than the sharpest expert; on the contrary, due to the product form~\eqref{eq_PoE}, the PoE can be sharper than any of the experts. This can be confirmed in Fig.~\ref{Fig_local_approxs}: the PoE produces poor prediction mean and overconfident prediction variance by aggregating the predictions from six independent experts, due to the inability of suppressing poor experts; on the contrary, the MoE provides desirable predictions through gating functions.

Hence, in order to improve PoE, we retain the effective training process but modify the predicting process. Instead of following the simple product rule to aggregate the experts' predictions, various \textit{aggregation} criteria have been proposed to weaken the votes of poor experts.\footnote{Note that the aggregation strategy is \textit{post-processing} or \textit{transductive} since it is independent from model training but depends on the test point location.} Particularly, the aggregations are expected to have several properties~\cite{cao2014generalized}: (i) the aggregated prediction is sensible in terms of probability, and (ii) the aggregated prediction is robust to weak experts.

Given the GP experts $\{ \mathcal{M}_i \}_{i=1}^M$ with predictive distributions $\{p(y_*|\mathcal{D}_i, \bm{x}_*) = \mathcal{N}(\mu_i(\bm{x}_*), \sigma^2_i(\bm{x}_*))\}_{i=1}^M$ at $\bm{x}_*$, the PoEs~\cite{hinton2002training, cao2014generalized, chen2009bagging, okadome2013fast} aggregate the experts' predictions through a modified product rule as
\begin{equation} \label{eq_PoE_post}
p(y_*|\mathcal{D}, \bm{x}_*) = \prod_{i=1}^M p_i^{\beta_{*i}}(y_*|\mathcal{D}_i, \bm{x}_*),
\end{equation}
where $\beta_{*i}$ is a weight quantifying the contribution of $p_i(y_*|\mathcal{D}_i, \bm{x}_*)$ at $\bm{x}_*$. Using~\eqref{eq_PoE_post}, we can derive the aggregated prediction mean and variance with closed-form expressions. The original product-rule aggregation~\cite{hinton2002training} employs a constant weight $\beta_{*i} = 1$, resulting in the aggregated precision $\sigma^{-2}(\bm{x}_*) = \sum_{i=1}^M \sigma^{-2}_i(\bm{x}_*)$ which will explode rapidly with increasing $M$. To alleviate the overconfident uncertainty, the generalized PoE (GPoE)~\cite{cao2014generalized} introduces a varying weight $\beta_{*i}$, which is defined as the difference in the differential entropy between the expert's prior and posterior, to increase or decrease the importance of experts based on their prediction uncertainty. But with this flexible weight, the GPoE produces explosive prediction variance when leaving the training data~\cite{liu2018generalized}. To address this issue, we can impose a constraint $\sum_{i=1}^M \beta_{*i} = 1$, see the favorable predictions in Fig.~\ref{Fig_local_approxs}; or we can employ a simple weight $\beta_{*i} = 1/M$ such that the GPoE recovers the GP prior when leaving $\bm{X}$, at the cost of however producing underconfident prediction variance~\cite{deisenroth2015distributed}.\footnote{With $\beta_{*i} = 1/M$, the GPoE's prediction mean is the same as that of PoE, but the prediction variance blows up as $M$ times that of PoE.}

Alternatively, Bayesian committee machine (BCM)~\cite{tresp2000bayesian, ng2014hierarchical, deisenroth2015distributed, mair2018distributed} aggregates the experts' predictions from another point of view by imposing a conditional independence assumption $p(\bm{y}|y_*) \approx \prod_{i=1}^M p(\bm{y}_i | y_*)$, which in turn explicitly introduces a \textit{common} prior $p(y_*|\bm{\theta})$ for experts.\footnote{The BCM can be interpreted as a sparse GP which treats the test inputs as inducing points~\cite{quinonero2005unifying}.} Thereafter, by using the Bayes rule we have
\begin{equation} \label{eq_BCM}
p(y_*|\mathcal{D}, \bm{x}_*, \bm{\theta}) = \frac{\prod_{i=1}^M p_i^{\beta_{*i}}(y_*|\mathcal{D}_i, \bm{x}_*, \bm{\theta})}{p^{\sum_{i=1}^M \beta_{*i} - 1}(y_*|\bm{\theta})}.
\end{equation}
The prior correlation term helps BCM recover the GP prior when leaving $\bm{X}$, and the varying $\beta_{*i}$ akin to that of GPoE helps produce robust BCM (RBCM) predictions within $\bm{X}$~\cite{deisenroth2015distributed}. The BCM however will produce unreliable prediction mean when leaving $\bm{X}$, which has been observed and analyzed in~\cite{deisenroth2015distributed, liu2018generalized}. Notably, unlike PoE, the common prior in BCM requires that all the experts should share the hyperparameters: that is why we explicitly write~\eqref{eq_BCM} conditioned on $\bm{\theta}$.

It has been pointed out that the conventional PoEs and BCMs are \textit{inconsistent}~\cite{liu2018generalized, szabo2017asymptotic}. That is, their aggregated predictions cannot recover that of full GP when $n \rightarrow \infty$. To raise consistent aggregation, the nested pointwise aggregation of experts (NPAE)~\cite{rulliere2018nested} removes the independence assumption by assuming that $\bm{y}_i$ has not yet been observed such that $\mu_i(\bm{x}_*)$ is a random variable. The NPAE provides theoretically consistent predictions at the cost of requiring much higher time complexity due to the inversion of a new $M \times M$ kernel matrix at each test point. To be efficient while retaining consistent predictions, instead of using the fixed GP prior to correct the aggregation like (R)BCM, the generalized RBCM (GRBCM)~\cite{liu2018generalized} (i) introduces a \textit{global} communication expert $\mathcal{M}_c$ to perform correction, i.e., acting as a base expert, and (ii) considers the covariance between global and local experts to improve predictions, leading to the aggregation
\begin{equation} \label{Eq_GRBCM}
\begin{aligned}
p(y_*|\mathcal{D},\bm{x}_*) &= \frac{\prod_{i=2}^{M} p_{+i}^{\beta_{*i}}(y_*|\mathcal{D}_{+i}, \bm{x}_*,)}{p_c^{\sum_{i=2}^M \beta_{*i} -1}(y_*|\mathcal{D}_c,\bm{x}_*)},
\end{aligned}
\end{equation}
where $p_c(y_*|\mathcal{D}_c,\bm{x}_*)$ is the predictive distribution of $\mathcal{M}_c$, and $p_{+i}(y_*|\mathcal{D}_{+i}, \bm{x}_*)$ is the predictive distribution of an enhanced expert $\mathcal{M}_{+i}$ trained on the augmented dataset $\mathcal{D}_{+i} = \{ \mathcal{D}_i, \mathcal{D}_c \}$. Different from (R)BCM, GRBCM employs the informative $p_c(y_*|\mathcal{D}_c,\bm{x}_*)$ rather than the simple prior to support consistent predictions when $n \to \infty$.

Note that the implementations of these transductive aggregations usually share the hyperparameters across experts~\cite{deisenroth2015distributed}, i.e., $\bm{\theta}_i = \bm{\theta}$, because (i) it achieves automatic regularization for guarding against local over-fitting, and eases the inference due to fewer hyperparameters; (ii) it allows to temporarily ignore the noise term of GP in aggregation, i.e., using $p_i(f_*|\mathcal{D}_i, \bm{x}_*)$ instead of $p_i(y_*|\mathcal{D}_i, \bm{x}_*)$ like~\cite{deisenroth2015distributed}, to relieve the inconsistency of typical aggregations; and finally (iii) the BCMs cannot support experts using individual hyperparameters as discussed before. But the shared hyperparameters limit the capability of capturing non-stationary features, which is the superiority of local approximations.\footnote{When sharing hyperparameters, the local structure itself may have good estimations of hyperparameter to capture some kind of local patterns~\cite{liu2019understanding}.} Besides, another main drawback of aggregations is the \textit{Kolmogorov inconsistency}~\cite{samo2016string} induced by the separation of training and predicting such that it is not a unifying probabilistic framework. That is, when we extend the predictive distributions at multiple test points, e.g., $\bm{x}_*$ and $\bm{x}'_*$, we have $p(y_{*}|\mathcal{D}) \ne \int p(y_{*},y'_{*}|\mathcal{D}) dy'_{*}$.

\section{Improvements over scalable GPs} \label{sec_global_local}
\subsection{Scalability}
The reviewed global approximations, especially the sparse approximations, have generally reduced the standard cubic complexity to $\mathcal{O}(nm^2)$ through $m$ inducing points. Moreover, their complexity can be further reduced through SVI~\cite{hensman2013gaussian} (with $\mathcal{O}(m^3)$) and the exploitation of structured data~\cite{wilson2015kernel} (with $\mathcal{O}(n + d\log m^{1+1/d})$). Sparse approximations however are still computationally impractical in the scenarios requiring real-time predictions, for example, environmental sensing and monitoring~\cite{low2011active}. Alternatively, we can implement sparse approximations using advanced computing infrastructure, e.g., Graphics Processing Units (GPU) and distributed clusters/processors, to further speed up the computation.

Actually, the exact GP using GPU and distributed clusters has been investigated~\cite{franey2012short, paciorek2015parallelizing, ambikasaran2016fast, tavassolipour2017learning, wang2019exact} in the regime of distributed learning~\cite{dean2008mapreduce}. The direct strategy implements parallel and fast linear algebra algorithms, e.g., the HODLR algorithm~\cite{ambikasaran2016fast} and the MVM algorithm, with modern distributed memory and multi-core/multi-GPU hardware.\footnote{Wang et al.~\cite{wang2019exact} successfully trained a MVM-based exact GP over a million data points in three days through eight GPUs.} 

In the meantime, the GPU accelerated sparse GPs have been explored. Since most of the terms in~\eqref{eq_F_q} can be factorized over data points, the inference can be parallelized and accelerated by GPU~\cite{gal2014distributed, dai2014gaussian}. Moreover, by further using the relaxed variational lower bound~\eqref{eq_F_Q} or grid inducing points in~\eqref{eq_SKI_Qnn}, the TensorFlow-based \textit{GPflow} library~\cite{matthews2017gpflow} and the PyTorch-based \textit{GPyTorch} library~\cite{gardner2018gpytorch} have been developed to exploit the usage of GPU hardwares. 

Besides, the parallel sparse GPs, e.g., the parallel PIT(C) and Incomplete Cholesky Factorization (ICF), have been developed using the message passing interface framework to distribute computations over multiple machines~\cite{chen2012decentralized, chen2013parallel}. Ideally, the parallelization can achieve a speed-up factor close to the number of machines in comparison to the centralized counterparts. Recently, a unifying framework which distributes conventional sparse GPs, including DTC, FI(T)C, PI(T)C and low-rank-cum-markov approximation (LMA)~\cite{low2015parallel}, have been built via varying correlated noise structure~\cite{hoang2016distributed}. Impressively, Peng et al.~\cite{peng2017asynchronous} first implemented the sparse GPs in a distributed computing platform using up to one \textit{billion} training points, and trained the model successfully within two hours. 

The local approximations generally have the same complexity to the global approximations if the training size $m_0$ for each expert is equal to the inducing size $m$. The local opinion however naturally encourages the parallel/distributed implementations to further reduce computational complexity, see for example~\cite{gramacy2016lagp, gramacy2016speeding, gramacy2014massively}.

\subsection{Capability}
\label{sec_model_capability}
Originated from the low-rank Nystr\"{o}m approximation, the global sparse approximations have been found to work well for approximating \textit{slow-varying} features with high spatial correlations. This is because in this case, the spectral expansion of the kernel matrix $\bm{K}_{nn}$ is dominated by a few large eigenvectors. On the contrary, when the latent function $f$ has \textit{quick-varying} (non-stationary) features, e.g., the complicated time series tasks~\cite{ding2017multiresolution}, the limited global inducing set struggles to exploit the local patterns. The D\&C inspired local approximations however are capable of capturing local patterns but suffer from the inability of describing global patterns. Hence, in order to enhance the representational capability of scalable GPs, the \textit{hybrid} approximations are a straightforward thread by combining global and local approximations in tandem. 

Alternatively, the hybrid approximations can be accomplished through an \textit{additive process}~\cite{snelson2007local,vanhatalo2010approximate, vanhatalo2008modelling}. For instance, after partitioning the input space into subregions, the partially independent conditional (PIC)~\cite{snelson2007local} and its stochastic and distributed variants~\cite{hoang2015unifying, hoang2016distributed, yu2017stochastic} extend PITC by retaining the conditional independence of training and test, i.e., $\bm{f} \perp f_* | \bm{f}_m$, for all the subregions except the one containing the test point $\bm{x}_*$, thus enabling the integration of local and global approximations in a \textit{transductive} pattern. Mathematically, suppose that $\bm{x}_* \in \Omega_j$ we have
\begin{equation}
q(\bm{f}, f_* | \bm{f}_m) = p(\bm{f}_j, f_*|\bm{f}_m) \prod_{i \neq j}^M p(\bm{f}_i | \bm{f}_m).
\end{equation}
This model corresponds to an exact GP with an \textit{additive} kernel
\begin{equation*} \label{eq_PIC_kernel}
k_{\mathrm{PIC}}(\bm{x}_i, \bm{x}_j) = k_{\mathrm{SoR}}(\bm{x}_i, \bm{x}_j) + \psi_{ij} [k(\bm{x}_i, \bm{x}_j) - k_{\mathrm{SoR}}(\bm{x}_i, \bm{x}_j)],
\end{equation*}
where $\psi_{ij} = 1$ when $\bm{x}_i$ and $\bm{x}_j$ belong to the same block; otherwise $\psi_{ij} = 0$. Note that the hybrid PIC recovers FIC by taking all the subregion sizes to one; it is left with the purely local GPs by taking the inducing size to zero. The additive kernel similar to $k_{\mathrm{PIC}}$ has also been employed in~\cite{vanhatalo2010approximate, vanhatalo2008modelling, gu2012spatial} by combining the CS kernel~\cite{gneiting2002compactly, wendland2004scattered} and the sparse approximation. Furthermore, as an extension of PIC, the tree-structured GP~\cite{bui2014tree} ignores most of the inter-subregion dependencies of inducing points, but concentrates on the dependency of adjacent subregions lying on a chained tree structure. The almost purely localized model reduces the time complexity to be linear to $n$ and allows using many inducing points. 

Hybrid approximations can also be conducted through a \textit{coarse-to-fine} process, resulting in a hierarchical structure with multiple layers yielding multi-resolution~\cite{lee2017hierarchically, park2010hierarchical, fox2012multiresolution, nychka2015multiresolution}. For example, Lee et al.~\cite{lee2017hierarchically} extended the work of~\cite{bui2014tree} into a hierarchically-partitioned GP approximation. This model has multiple layers, with the root layer being localized GPs. Particularly, each layer owns the individual kernel, the configuration of which is determined by the density of inducing points; the adjacent layers share a cross-covariance function which is convolved from two relevant kernels, like~\cite{alvarez2011computationally}. Similarly, Park et al.~\cite{park2010hierarchical} presented a two-layer model, wherein a GP is placed over the centroids of the subsets as $g(\bm{c}) \sim \mathcal{GP}(0, k_g(\bm{c}, \bm{c}'))$ to construct a rough global approximation in the top layer; then in each subregion of the root layer, a local GP is trained by using the global-level GP as the mean prior $f_i(\bm{x}) \sim \mathcal{GP}(g(\bm{x}), k_i(\bm{x}, \bm{x}'))$. This model has also been improved into multi-layer structure~\cite{fox2012multiresolution}.

Inevitably, the combination of local approximations may induce discontinuous predictions and inaccurate uncertainties on the boundaries of subregions. For example, the tree-structured GPs~\cite{bui2014tree,lee2017hierarchically} completely adopt a localized predictive distribution, which suffers from severe discontinuity. The predictions could be smoothed by placing inducing points on the boundaries of subregions~\cite{lee2017hierarchically}, which however is hard to implement. The PIC predictive distribution is composed of both global and local terms~\cite{snelson2007local}, which partially alleviates the discontinuity. To completely address the discontinuity, Nguyen et al.~\cite{nguyen2014fast, nguyen2016variational} combined sparse approximations with model averaging strategies, e.g., MoE.

Finally, despite the hybrid approximations, the representational capability of sparse approximations can be enhanced through a more powerful probabilistic framework. For instance, the inter-domain GP~\cite{lazaro2009inter}, which employs an idea similar to the convolution process in multi-output GP~\cite{alvarez2011computationally, liu2018remarks} and high-dimensional GP~\cite{van2017convolutional}, uses a linear integral transform $g(\bm{z}) = \int w(\bm{x}, \bm{z}) f(\bm{x}) d\bm{x}$ to map the inducing points into another domain of possibly different dimensions.\footnote{$g(\bm{z})$ is still a GP by a linear transform of $f(\bm{x})$. Besides, with $w(\bm{x}, \bm{z}) = \delta(\bm{x} - \bm{z})$ where $\delta$ is a Dirac delta, the inter-domain GP recovers FITC.} The inducing variables in the new domain can own a new kernel and induce richer dependencies in the old domain. The inter-domain idea has also been applied to the posterior approximations~\cite{tobar2015learning, matthews2016sparse, bui2017unifying}. Besides, from the weight-space view in~\eqref{eq_weight_sparse}, it is encouraged to employ different configurations for the basis functions to capture slow- and quick-varying features using different scales~\cite{walder2008sparse, zhang2015efficient}. This kind of weight-space non-stationary GP indeed can be derived from the inter-domain view, see~\cite{lazaro2009inter}.

Alternatively, unlike the standard GP using a homoscedastic noise $\epsilon \sim \mathcal{N}(0, \sigma^2_{\epsilon})$, the FITC has been extended by a varying noise as $p(\bm{\epsilon}) = \mathcal{N}(\bm{\epsilon}|\bm{0}, \mathrm{diag}[\bm{h}])$ where $\bm{h} = [\sigma^2_{\epsilon}(\bm{x}_1), \cdots, \sigma^2_{\epsilon}(\bm{x}_n)]^{\mathsf{T}}$~\cite{snelson2006variable}. Moreover, Hoang et al.~\cite{hoang2016distributed} employed a $B$-th order Markov property on the correlated noise process $p(\bm{\epsilon}) = \mathcal{N}(\bm{\epsilon}|\bm{0}, \bm{K}_{\epsilon})$ in a distributed variational framework. The unifying framework accommodates existing sparse approximations, e.g., DTC and PIC, by varying the Markov order and noise structure.\footnote{It achieves the state-of-the-art results on the airline dataset with up to two million data points.} Yu et al.~\cite{yu2017stochastic} further extended this work through Bayesian treatment of hyperparameters to guard against over-fitting. More elegantly, Almosallam et al.~\cite{almosallam2016gpz} derived a scalable heteroscedastic Bayesian model from the weight-space view by adopting an additional log GP, which is analogous to~\cite{goldberg1998regression, lazaro2011variational}, to account for noise variance as $\sigma^2_{\epsilon}(\bm{x}_i) = \exp (\bm{\phi} (\bm{x}_i) \bm{w} + b)$. Differently, Liu et al.~\cite{liu2018large} derived the stochastic and distributed variants of~\cite{goldberg1998regression} for scalable heteroscedastic regression. They found that the distributed variant using experts with hybrid parameters improves both scalability and capability, while the stochastic variant using global inducing set may sacrifice the prediction mean for describing the heteroscedastic noise.

\section{Extensions and open issues}
\label{sec_futurework}
\subsection{Scalable manifold GP}
In scalable GP literature, we usually focus on the scenario wherein the training size $n$ is large (e.g., $n \ge 10^4$) whereas the number $d$ of inputs is modest (e.g., up to hundreds of dimensions). However, in practice we may need to handle the task with comparable $n$ and $d$ or even $d \gg n$, leading to the demand of high-dimensional scalable GP. 
In practice, we often impose low-dimensional constraints to restrict the high-dimensional problems, i.e., the inputs often lie in a \textit{p}-dimensional ($p < d$) manifold embedded in the original \textit{d}-dimensional space. This is because high-dimensional statistical inference is solvable only when the input size $d$ is compatible with the statistical power based on the training size $n$~\cite{yang2016bayesian}.

Hence, various manifold GPs~\cite{lawrence2005probabilistic, snelson2006variable}, which are expressed as
\begin{equation}
y = f(\bm{\Upsilon} \bm{x}) + \epsilon,
\end{equation}
where $\bm{\Upsilon} \in R^{p \times d}$ is a mapping matrix, have been developed to tackle high-dimensional big data through linear/nonlinear dimensionality reduction~\cite{snelson2006variable, chen2010compressive, damianou2014variational} or neural network-like input transformation~\cite{calandra2016manifold}. As a result, the kernel operates the data in a lower dimensional space as $k(\bm{\Upsilon} \bm{x}, \bm{\Upsilon} \bm{x}')$. Note that the mapping matrix and the scalable regression are learned jointly in the Bayesian framework for producing favorable results. Particularly, the true dimensionality of the manifold can be estimated using Bayesian mixture models~\cite{reich2011sufficient}, which however induce a heavy computational budget.

A recent exciting theoretical finding~\cite{yang2016bayesian} turns out that the learning of the intrinsic manifold can be bypassed, since the GP learned in the original high-dimensional space can achieve the optimal rate when $f$ is not highly smooth. This motivates the use of Bayesian model averaging based on random compression over various configurations in order to reduce computational demands~\cite{guhaniyogi2016compressed}.

Continual theoretical and empirical efforts are required for designing specific components, e.g., the convolutional kernel~\cite{van2017convolutional}, for scalable manifold GPs, because of the urgent demands in various fields, e.g., computer vision (CV).

\subsection{Scalable deep GP}
Motivated by the enormous success of deep learning in various fields, the scalable deep GPs~\cite{wilson2016deep, damianou2013deep} have been investigated in recent years.\footnote{GP has been pointed out as a shallow but infinitely wide neural network with Gaussian weights~\cite{neal1996bayesian, matthews2018gaussian}.}~A simple representative is combining the structural NNs and the flexible non-parametric GP together, wherein NNs map the original input space to the feature space for extracting non-stationary/recurrent features, and the last-layer sparse GP conducts standard regression over the latent space~\cite{wilson2016deep, wilson2016stochastic, calandra2016manifold, al2017learning}. The parameters of NNs and GP are jointly learned by maximizing the marginal likelihood in order to guard against over-fitting. The NNs+GP structure produces sensible uncertainties, and is found to be robust to adversarial examples in CV tasks~\cite{bradshaw2017adversarial}. Particularly, Cremanns and Roos~\cite{cremanns2017deep} employed the same hybrid structure, but used the NNs to learn input-dependent hyperparameters for the additive kernels. Then, the NeNe algorithm is employed to ease the GP inference. Besides, Iwata and Ghahramani~\cite{iwata2017improving} used the outputs of NNs as prior mean for SVGP~\cite{hensman2013gaussian}. 

More elegantly, inspired by deep learning, the deep GP (DGP)~\cite{damianou2013deep} and its variants~\cite{dai2015variational, bui2016deep, salimbeni2017doubly, cutajar2017random, havasi2018inference}, which employ the hierarchical and functional composite
\begin{equation}
y(\bm{x}) = f_l(f_{l-1}(\cdots f_1(\bm{x}))) + \epsilon
\end{equation}
to stack multiple layers of latent variable model (LVM)~\cite{lawrence2005probabilistic} for extracting features. The DGP showcases great flexibility in (un)supervised scenarios, resulting in however a non-standard GP. The recently developed convolutional kernel~\cite{van2017convolutional} opens up the way of DGP for CV tasks~\cite{kumar2018deep}. Note that the inference in DGP is intractable and expensive, thus efficient training requires a sophisticated approximate inference via inducing points~\cite{damianou2016variational, salimbeni2017doubly}, which in turn may limit the capability. Easier inference without loss of prediction accuracy has always been a big challenge for DGP to completely show its potential beyond regression.

\subsection{Scalable multi-task GP}
Due to the multi-task problems that have arose in various fields, e.g., environmental sensor networks and structure design, multi-task GP (MTGP)~\cite{alvarez2012kernels, liu2018remarks}, also known as multi-output GP, seeks to learn the latent $T$ correlated tasks $\bm{f} = [f_1, \cdots, f_T]^{\mathsf{T}}: R^d \mapsto R^T$ simultaneously as
\begin{equation}
\bm{f}(\bm{x}) \sim \mathcal{GP}(\bm{0}, \bm{k}_{\mathrm{MTGP}}(\bm{x},\bm{x}')), \quad \bm{y}(\bm{x}) = \bm{f}(\bm{x}) + \bm{\epsilon},
\end{equation}
where $\bm{\epsilon} = [\epsilon_1, \cdots, \epsilon_T]^{\mathsf{T}}$ is the individual noises. The crucial in MTGP is the construction of a valid multi-task kernel $\bm{k}_{\mathrm{MTGP}}(\bm{x},\bm{x}') \in R^{T \times T}$, which can be built through for example linear model of coregionalization~\cite{bonilla2008multi} and convolution process~\cite{alvarez2011computationally}. Compared to individual modeling of tasks which loses valuable information, the joint learning of tasks enables boosting predictions by exploiting the task correlations and leveraging information across tasks. 

Given that each of the $T$ tasks has $n$ training points, MTGP collects the data from all the tasks and fuses them in an entire kernel matrix, leading to a much higher complexity $\mathcal{O}(T^3n^3)$. Hence, since the inference in most MTGPs follows the standard process, the above reviewed sparse approximations and local approximations have been applied to MTGPs~\cite{alvarez2011computationally, nguyen2014collaborative, zhao2016variational, chiplunkar2016approximate} to improve the scalability.

To date, scalable MTGPs are mainly studied in the scenario where the tasks have well defined labels and share the input space with modest dimensions. Many efforts are required for extending current MTGPs to handle the 4V challenges in the regime of multi-task (multi-output) learning~\cite{xu2019survey}.

\subsection{Scalable online GP}
Typical it is assumed that the entire data $\mathcal{D}$ is available a priori to conduct the \textit{off-line} training. We however should consider the scenario where the data arrives sequentially, i.e., \textit{online} or \textit{streaming} data, in small unknown batches. For the complicated online regression, the model~\cite{csato2002sparse} should (i) have real-time adaptation to the streaming data; and (ii) handle large-scale case since the new data is continuously arriving.

Sparse GPs are extensible for online learning since they employ a small inducing set to summarize the whole training data~\cite{petelin2014evolving, huber2014recursive, le2017gogp}. As a result, the arrived new data interacts only with the inducing points to enhance fast online learning. This is reasonable since the updates of $\mu(\bm{x}_*)$ and $\sigma^2(\bm{x}_*)$ of FITC and PITC only rely on the inducing set and new data~\cite{bijl2015online, bijl2017system}. Moreover, the stochastic variants naturally showcase the online structure~\cite{cheng2016incremental}, since the bound in~\eqref{eq_F_Q} supports mini-batch learning by stochastic optimization. 

But there are two issues for scalable online GPs. First, some of them~\cite{huber2014recursive, bijl2015online, bijl2017system} fix the hyperparameters to obtain constant complexity per update. It is argued that empirically the optimization improves the model significantly in the first few iterations~\cite{cheng2016incremental}. Hence, with the advanced computing power and the demand of accurate predictions, it could update hyperparameters online over a small number of iterations. 

Second, the scalable online GPs implicitly assume that the new data and old data are drawn from the same input distribution. This however is not the case in tasks with complex trajectories, e.g., an evolving time-series~\cite{bui2017streaming}. To address the \textit{evolving} online learning, Nguyen et al.~\cite{nguyen2009local} presented a simple and intuitive idea using local approximations. This method maintains multiple local GPs, and either uses the new data to update the specific GP when they fall into the relevant local region, or uses the new data to train a new local GP when they are far away from the old data, resulting in however information loss from available training data. As the extension of~\cite{csato2002sparse, xu2014gp}, Bui et al.~\cite{bui2017streaming} deployed an elegant probabilistic framework to update the posterior distributions and hyperparameters in an online fashion, where the interaction happens between the old and new inducing points. The primary theoretical bounds for this Bayesian online inference model were also provided in~\cite{nguyen2017online}.

\subsection{Scalable recurrent GP}
There exist various tasks, e.g., speech recognition, system identification, energy forecasting and robotics, wherein the datasets are sequential and the ordering matters. Here, we focus on recurrent GP~\cite{kocijan2005dynamic, frigola2014variational} to handle sequential data.

The popular recurrent GP is GP-based nonlinear auto-regressive models with exogenous inputs (GP-NARX)~\cite{kocijan2005dynamic}, which is generally expressed as
\begin{equation}
\bm{x}_t = [y_{t-1}, \cdots, y_{t-L_y}, u_{t-1}, \cdots, u_{t-L_u}], \quad y_t = f(\bm{x}_t) + \epsilon_{y_t},
\end{equation}
where $u_t$ is the external input, $y_t$ is the output observation at time $t$, $L_y$ and $L_u$ are the lagged parameters that respectively indicate the numbers of delayed outputs and inputs to form the regression vector $\bm{x}_t$, $f$ is the emission function, and $\epsilon_{y_t}$ accounts for observation noise. Note that the observations $y_{t-1}, \cdots, y_{t-L_y}$ here are considered to be deterministic. The transformed input $\bm{x}_t$ comprising previous observations and external inputs enables using standard scalable GPs, e.g., sparse approximations, to train the GP-NARX. Due to the simplicity and applicability, GP-NARX has been well studied and extended to achieve robust predictions against outliers~\cite{mattos2015empirical}, local modeling by incorporating prior information~\cite{avzman2011dynamical}, and higher-order frequency response functions~\cite{worden2014gaussian}. A main drawback of GP-NARX however is that it cannot account for the observation noise in $\bm{x}_t$, leading to the errors-in-variables problem. To address this issue, we could (i) conduct data-preprocessing to remove the noise from data~\cite{frigola2013integrated}; (ii) adopt GPs considering input noise~\cite{damianou2016variational}; and (iii) employ the more powerful state space models (SSM)~\cite{frigola2014variational} introduced below.

The GP-SSM employs a more general recurrent structure as
\begin{equation}
\bm{x}_t = g(\bm{x}_{t-1}, u_{t-1}) + \epsilon_{x_t}, \quad \bm{y}_t = f(\bm{x}_{t}) + \epsilon_{y_t},
\end{equation}
where $\bm{x}_t$ is the state of the system which acts as an internal memory, $g$ is the transition function, $f$ is the emission function, $\epsilon_{x_t}$ is the transition noise, and finally $\epsilon_{y_t}$ is the emission noise. Note that GP-NARX is a simplified GP-SSM model with observable state. The GP-SSM takes into account the transition noise and brings the flexibility in requiring no lagged parameters. But this model suffers from intractable inference since we need to marginalize out all the latent variables, thus requiring approximate inference~\cite{frigola2014variational, frigola2013bayesian, svensson2016computationally}.

Finally, we again see the trend in combining recurrent GPs with neural networks. For instance, the deep recurrent GP~\cite{mattos2017deep} attempts to mimic the well-known recurrent neural networks (RNNs), with each layer modeled by a GP. Similar to DGP~\cite{damianou2013deep}, the inference in deep recurrent GP is intractable and requires sophisticated approximations, like~\cite{mattos2019stochastic}. Hence, to keep the model as simple as possible while retaining the recurrent capability, the long short-term memory (LSTM) model~\cite{hochreiter1997long} is combined with scalable GPs, resulting in analytical inference and desirable results~\cite{al2017learning}.

\subsection{Scalable GP classification}
Different from the regression tasks mainly reviewed by this article with continuous real observations, the classification has discrete class labels. To this end, the binary GP classification (GPC) model~\cite{nickisch2008approximations} with $y \in \{0, 1\}$ is usually formulated as
\begin{equation}
\label{GPC_binary}
f \sim \mathcal{GP}(0, k), \quad p(y|f) = \mathrm{Bernoulli}(\pi(f)),
\end{equation}
where $\pi(.) \in [0, 1]$ is an inverse link function\footnote{The conventional inverse link functions include the step function, the (multinomial) probit/logit function, and the softmax function.} that squashes $f$ into the class probability space. Differently, the multi-class GPC (MGPC)~\cite{kim2006bayesian} with $y \in \{1, \cdots, C\}$ is
\begin{equation}
\label{GPC_multiclass}
f^c \sim \mathcal{GP}(0, k_c),  \quad p(y|\bm{f}) = \mathrm{Categorical}(\pi(\bm{f})),
\end{equation}
where $\{f^c\}_{c=1}^C$ are independent latent functions\footnote{In contrast, inspired by the idea of MTGP, the correlations among latent functions have been exploited by~\cite{chai2012variational}.} for $C$ classes, and $\bm{f} = [f^1, \cdots, f^C]^{\mathsf{T}}: R^d \mapsto R^C$.\footnote{Similar to GPR, potential classification error can be considered through $f = \hat{f} + \epsilon$. By varying over different noise distributions, e.g., Gaussian and logistic, we however could recover conventional inverse link functions, e.g., probit and logit functions, through the integration of $\epsilon$~\cite{ruiz2018augment}.} Due to the non-Gaussian likelihood, exact inference for GPC however is intractable, thus requiring approximate inference, the key of which approximates the non-Gaussian posterior $p(f|y) \propto p(y|f)p(f)$ with a Gaussian $q(f|y)$~\cite{nickisch2008approximations}.

Motivated by the success of scalable GPR, we could directly treat GPC as a regression task~\cite{frohlich2013large}; or solve it as GPR by a transformation that interprets class labels as outputs of a Dirichlet distribution~\cite{milios2018dirichlet}. This sidesteps the non-Gaussian likelihood. A more principled way however is adopting GPCs in~\eqref{GPC_binary} and~\eqref{GPC_multiclass}, and combining approximate inference, e.g., laplace approximation, EP and VI, with the sparse strategies in section~\ref{sec_sparse_approximations} to derive scalable GPCs~\cite{naish2008generalized, hensman2015mcmc, hensman2015scalable, hernandez2016scalable, villacampa2017scalable, wenzel2018efficient}.\footnote{Different from sparse GPR, the inducing points optimized in sparse GPC are usually pushed towards decision boundary~\cite{hensman2015scalable}.}

The main challenges of scalable GPC, especially MGPC, are: (i) the intractable inference and posterior, and (ii) the high training complexity for a large $C$. For the first issue, the stochastic GPC derives the model evidence expressed as the integration over an one-dimensional Gaussian distribution, which can be adequately calculated using Gaussian-Hermite quadrature~\cite{hensman2015scalable, matthews2017scalable}. Furthermore, the GPC equipped with the FITC assumption owns a completely analytical model evidence~\cite{hernandez2016scalable, villacampa2017scalable}. Particularly, when taking the logit/softmax inverse link function, the P\`olya-Gamma data augmentation~\cite{polson2013bayesian} offers analytical inference and posterior for GPC~\cite{wenzel2018efficient, Galy2018efficient}. For the second issue, since the complexity of MGPC is linear to the number $C$ of classes, alternatively, we may formulate the model evidence as a sum over classes like~\cite{ruiz2018augment}, thus allowing efficient stochastic training.

\section{Conclusions}
\label{sec_conclusions}
Although the GP itself has a long history, the non-parametric flexibility and the high interpretability make it popular yet posing many challenges in the era of big data. In this paper, we have attempted to summarize the state of scalable GPs in order to (i) well understand the state-of-the-art, and (ii) attain insights into new problems and discoveries. The extensive review seeks to uncover the applicability of scalable GPs to real-world large-scale tasks, which in turn present new challenges, models and theory in the GP community.

\appendices
\section{Libraries and datasets}
\begin{table*}
	\caption{List of primary libraries supporting representative scalable GPs.}
	\label{Tab_GP_libraries}
	\centering
	\resizebox{\textwidth}{!}{
		\begin{tabular}{lcccccc}
			\hline
			Package & Language & Global & Local & Others & GPU supported \\\hline
			GPML & Matlab & FITC~\cite{snelson2006sparse}, VFE~\cite{titsias2009variational}, SPEP~\cite{bui2017unifying}, SKI~\cite{wilson2015kernel} & - & - & - \\
			GPy & Python & VFE~\cite{titsias2009variational}, SPEP~\cite{bui2017unifying},
			SKI~\cite{wilson2015kernel}, SVGP~\cite{hensman2013gaussian} & - & - & $\checkmark$ \\
			GPstuff & Matlab\&R & SoR~\cite{smola2001sparse}, DTC~\cite{csato2002sparse}, FITC~\cite{snelson2006sparse}, VFE~\cite{titsias2009variational}, SVGP~\cite{hensman2013gaussian}, CS~\cite{gneiting2002compactly} & - & PIC~\cite{snelson2007local}, CS+FIC~\cite{vanhatalo2008modelling} & - \\
			GPflow & Python & VFE~\cite{titsias2009variational}, SVGP~\cite{hensman2013gaussian} & - & NNs+SVGP~\cite{bradshaw2017adversarial} & $\checkmark$ \\
			pyMC3 & Python & DTC~\cite{csato2002sparse}, FITC~\cite{snelson2006sparse}, VFE~\cite{titsias2009variational} & - & - & - \\
			GPyTorch & Python & SKI~\cite{wilson2015kernel} & - & DKL (NNs+SKI)~\cite{wilson2016deep} & $\checkmark$ \\
			pyGPs & Python & FITC~\cite{snelson2006sparse} & - &- & - \\
			AugGP & Julia & VFE~\cite{titsias2009variational}, SVGP~\cite{hensman2013gaussian} & - & - & - \\
			laGP & R & - & NeNe~\cite{urtasun2008sparse} & - & $\checkmark$ \\
			GPLP & Matlab & - & NeNe~\cite{urtasun2008sparse}, PoE~\cite{chen2009bagging}, DDM~\cite{park2011domain} & PIC~\cite{snelson2007local} & - \\
			\hline
	\end{tabular}}
\end{table*}

Table~\ref{Tab_GP_libraries} summarizes the primary libraries that implement representative scale GPs and are well-known for both academia and industry.\footnote{The primary libraries include: (1) GPML (\url{http://www.gaussianprocess.org/gpml/});
(2) GPy (\url{https://github.com/SheffieldML/GPy});
(3) GPstuff (\url{https://github.com/gpstuff-dev/gpstuff});
(4) GPflow (\url{https://github.com/GPflow/GPflow});
(5) pyMC3 (\url{https://github.com/pymc-devs/pymc3});
(6) GPyTorch (\url{https://github.com/cornellius-gp/gpytorch});
(7) pyGPs (\url{https://github.com/PMBio/pygp});
(8) AugGP (\url{https://github.com/theogf/AugmentedGaussianProcesses.jl});
(9) laGP (\url{http://bobby.gramacy.com/r_packages/laGP/});
(10) GPLP (\url{http://www.jmlr.org/mloss/}).
}    
It is observed that Python and hardware acceleration are becoming popular for the GP community. Note that some specific scalable GP packages implementing the advanced models reviewed in sections~\ref{sec_global_local} and~\ref{sec_futurework} are not listed here, which can be found in the relevant researchers' webpage.

Besides, except the well-known UCI\footnote{\url{https://archive.ics.uci.edu/ml/index.php}} regression datasets with $n \in [15, 4.18\times10^6]$ and $d \in [3, 4.8\times10^5]$, and the LIBSVM\footnote{\url{https://www.csie.ntu.edu.tw/~cjlin/libsvm/}} regression datasets with  $n \in [152, 2.06\times10^4]$ and $d \in [6, 4.27\times10^6]$, Table~\ref{Tab_datasets} summarizes the regression datasets ($n \ge 10^4$) occurred in scalable GP literature. It is found that researchers have assessed scalable GPs with up to about one billion training points~\cite{peng2017asynchronous}.

\begin{table}
	\caption{Big regression datasets ($n \ge 10^4$) in the literature.}
	\label{Tab_datasets}
	\centering
	\begin{tabular}{lcc}
		\hline
		Dataset & No. of inputs & No. of observations \\\hline
		terrain~\cite{lee2017hierarchically} & 2 & 40.00K \\
		aimpeak~\cite{chen2013parallel} & 5 & 41.85K \\
		sarcos~\cite{rasmussen2006gaussian} & 21 & 48.93K \\
		natural sound~\cite{wilson2015kernel} & 1 & 59.31K \\
		chem~\cite{chalupka2013framework} & 15 & 63.07K \\
		kuka~\cite{meier2014incremental} & 28 & 197.92K \\
		crime~\cite{flaxman2015fast} & 2 & 233.09K \\
		sdss~\cite{almosallam2016gpz} & 10 & 300.00K \\
		precipitation~\cite{dong2017scalable} & 3 & 628.47K \\
		mujoco~\cite{cheng2017variational} & 23 & 936.35K \\
		airline~\cite{hensman2013gaussian} & 8 & 5.93M \\ 
		bimbo~\cite{rivera2017forecasting} & 147 & 9.05M \\ fortune~\cite{rivera2017forecasting} & 112 & 10.35M \\
		NYC taxi~\cite{peng2017asynchronous} & 9 & 1.21B \\
		\hline
	\end{tabular}
\end{table}

\section*{Acknowledgements}
This work was conducted within the Rolls-Royce@NTU Corporate Lab with support from the National Research Foundation (NRF) Singapore under the Corp Lab@University Scheme. It is also partially supported by the Data Science and Artificial Intelligence Research Center (DSAIR) and the School of Computer Science and Engineering at Nanyang Technological University.

\ifCLASSOPTIONcaptionsoff
  \newpage
\fi



%\bibliographystyle{IEEEtran}
%\begin{thebibliography}{1}

\bibliographystyle{IEEEtran}
\bibliography{IEEEabrv,bigGPSurvey}

%\end{thebibliography}



%\vfill


\end{document}